\documentclass[lettersize,journal,hidelinks]{IEEEtran}
\usepackage{amsmath,amssymb,amsfonts}
\usepackage{algorithm}
\usepackage{algorithmicx}
\usepackage[noend]{algpseudocode}
\usepackage{derivative}
\usepackage{graphicx}
\usepackage{pgfplots}
\pgfplotsset{compat=newest}
\usepackage{textcomp}
\usepackage{xcolor}
\usepackage{enumitem}
\usepackage{csquotes}
\usepackage{multirow}
\usepackage{booktabs}
\usepackage{colortbl}
\usepackage{tikz}
\definecolor{Gray}{gray}{0.94}
\usepackage{orcidlink}

\usepackage{gnuplottex}
\usepackage{xparse}

\def\nobreakbefore{%
  \relax\ifvmode\else
    \ifhmode
      \ifdim\lastskip > 0pt\relax
        \unskip\nobreakspace
      \fi
    \fi
  \fi
}
\let\oldcite\cite
\renewcommand\cite{\nobreakbefore\oldcite}

\usepackage{xspace}
\xspaceaddexceptions{)}
\usepackage{cite}
\usepackage[acronyms]{glossaries}

\newglossaryentry{vdd}{
	name={\ensuremath{V_{\text{DD}}}},
	description={supply voltage},
	symbol=Vdd,
	sort=vdd}
\newglossaryentry{vth}{
	name={\ensuremath{V_{\text{TH}}}},
	description={threshold voltage},
	symbol=Vth,
	sort=vth}
\newglossaryentry{gnd}{
	name={GND},
	description={El. ground},
	symbol=GND}

\newacronym{aef}{AEF}{average electric field}
\newacronym{dnn}{DNN}{deep neural network}
\newacronym{nn}{NN}{Neural network}
\newacronym{kan}{KAN}{Kolmogorov-Arnold network}
\newacronym{fkan}{FKAN}{Fourier KAN}
\newacronym{kart}{KART}{Kolmogorov-Arnold representation theorem}
\newacronym{cpn}{CPN}{current/charge prediction network}
\newacronym{pgn}{PGN}{parameter generator network}
\newacronym{gnn}{GNN}{graph neural network}
\newacronym{rnn}{RNN}{reccurrent neural network}
\newacronym{cnn}{CNN}{convolutional neural network}
\newacronym{aspp}{ASPP}{atrous spatial pyramid pooling}
\newacronym{mape}{MAPE}{mean absolute percentage error}
\newacronym{dtco}{DTCO}{design technology co-optimization}
\newacronym{iou}{IoU}{intersection over union}
\newacronym{fom}{FoM}{figure of merit}
\newacronym{miou}{MIoU}{mean intersection over union}
\newacronym{espp}{EsPP}{electrostatic potential profile}
\newacronym{mse}{MSE}{mean square error}
\newacronym{bjt}{BJT}{bipolar junction transistor}
\newacronym{ml}{ML}{Machine learning}
\newacronym{mlp}{MLP}{multi-layer perceptron}

\newacronym{rnet}{R-Net}{resistor network}

\newacronym{tcad}{TCAD}{Technology CAD}
\newacronym{cad}{CAD}{computer-aided design}
\newacronym{spice}{SPICE}{Simulation Program with Integrated Circuit Emphasis}
\newacronym{hspice}{HSPICE}{PrimeSim HSPICE}
\newacronym{ddic}{DDIC}{display driver integration circuit}
\newacronym{ss}{SS}{sub-threshold slope}

\newacronym{fdsoi}{FDSOI}{Fully-Depleted Silicon on Insulator}
\newacronym{fefet}{FeFET}{Ferroelectric FET}
\newacronym{finfet}{FinFET}{fin field-effect transistor}
\newacronym{fe}{FE}{ferroelectric}
\newacronym{mlc}{MLC}{multi-level cell}

\newacronym{nls}{NLS}{nucleation-limited switching}
\newacronym{ds}{DS}{Domain Stochasticity}

\newacronym{dtd}{DTD}{Device-to-Device}
\newacronym{ctc}{CTC}{Cycle-to-Cycle}
\newacronym{mac}{MAC}{multiply-accumulate}

\newacronym{stt}{STT}{Spin Transfer Torque}
\newacronym{rram}{ReRAM}{Resistive RAM}
\newacronym{pde}{PDE}{partial differential equation}

\newcommand{\nn}{\gls{nn}\xspace}

\newcommand{\kan}{\gls{kan}\xspace}
\newcommand{\fkan}{\gls{fkan}\xspace}
\newcommand{\kart}{\gls{kart}\xspace}

\newcommand{\mape}{\gls{mape}\xspace}
\newcommand{\dtco}{\gls{dtco}\xspace}

\newcommand{\mlp}{\gls{mlp}\xspace}

\newcommand{\hspice}{\gls{hspice}\xspace}

\newcommand{\finfet}{\gls{finfet}\xspace}

\newcommand{\vds}{\ensuremath{V_{\text{D}}}\xspace}
\newcommand{\vgs}{\ensuremath{V_{\text{G}}}\xspace}
\newcommand{\qd}{\ensuremath{Q_{\text{D}}}\xspace}
\newcommand{\qs}{\ensuremath{Q_{\text{S}}}\xspace}
\newcommand{\qg}{\ensuremath{Q_{\text{G}}}\xspace}
\newcommand{\qb}{\ensuremath{Q_{\text{B}}}\xspace}

\newcommand{\ids}{\ensuremath{I_{\text{D}}}\xspace}

\usepackage{array}
\newcolumntype{?}{!{\vrule width 1pt}}

\usepackage{flushend}

\usepackage{siunitx}
\DeclareSIUnit{\million}{\text{million}}
    
\usepackage{authblk}

\usepackage{lipsum}

\usepackage[skip=0pt]{caption}
\usepackage{subcaption}
\usepackage{gensymb}
\begin{document}


\title{Kolmogorov-Arnold Network for Transistor Compact Modeling}

\author{Rodion~Novkin$^{\orcidlink{0009-0006-6632-9804}}$, and Hussam~Amrouch$^{\orcidlink{0000-0002-5649-3102}}$,~\IEEEmembership{Member,~IEEE}

\thanks{
Rodion Novkin and Hussam Amrouch are with the Technical University of Munich; TUM School of Computation, Information and Technology;\\ Chair of AI Processor Design; Munich Institute of Robotics and Machine Intelligence, Munich, Germany (e-mail: \nolinkurl{{rodion.novkin,amrouch}@tum.de}).
}

\thanks{(\emph{Corresponding author: Rodion Novkin})}
}

\markboth{}
{Shell \MakeLowercase{\textit{Novkin et al.}}: Kolmogorov-Arnold Network for Transistor Compact Modeling}


\maketitle


\begin{abstract}
\nn-based transistor compact modeling has recently emerged as a transformative solution for accelerating device modeling and SPICE circuit simulations, particularly at advanced technology nodes, where the development cycle for mature physics-based models becomes very extensive. However, conventional \nn architectures, despite their widespread adoption in state-of-the-art methods, primarily function as black-box problem solvers. This lack of interpretability significantly limits their capacity to extract and convey meaningful insights into learned data patterns, posing a major barrier to their broader adoption in critical modeling tasks.
This work introduces, for the first time, \kan for the transistor - a groundbreaking \nn architecture that seamlessly integrates interpretability with high precision in physics-based function modeling. We systematically evaluate the performance of \kan and \fkan for \finfet compact modeling, benchmarking them against the golden industry-standard compact model and the widely used \mlp architecture. Our results reveal that \kan and \fkan consistently achieve superior prediction accuracy for critical figures of merit, including gate current (\ids), drain charge (\qd), and source charge (\qs). Furthermore, we demonstrate and improve the unique ability of \kan to derive symbolic formulas from learned data patterns - a capability that not only enhances interpretability but also facilitates in-depth transistor analysis and optimization.
This work highlights the transformative potential of \kan in bridging the gap between interpretability and precision in \nn-driven transistor compact modeling. By providing a robust and transparent approach to transistor modeling, \kan represents a pivotal advancement for the semiconductor industry as it navigates the challenges of advanced technology scaling.
\end{abstract}

\begin{IEEEkeywords}
Kolmogorov-Arnold Network, Transistor Compact Model, Symbolic Regression, Circuit Simulation
\end{IEEEkeywords}

\section{Introduction} \label{sec:intro}
 
\IEEEPARstart{T}{he} semiconductor industry relies heavily on accurate transistor models, which are essential for design optimization, variability analysis, and circuit simulation. The Berkeley Short-Channel IGFET Model (BSIM) \cite{Khandelwal12, Duarte2015, BSIMweb} is among the most widely used compact models, leveraging physics-based equations to accurately describe transistor behavior. While such analytical solutions remain the gold standard, their development demands extensive domain expertise, substantial time investment, and considerable effort. The complexity of underlying physics continues to increase with advancing technology nodes, introducing phenomena such as quantum effects and other intricate device behaviors. \nn-based transistor models have emerged as a promising alternative to accelerate the transistor modeling process by circumventing the need to delve deeply into underlying physics. The inherent capability of \nn to approximate complex data patterns enables a black-box approach to transistor modeling, where specific device inputs (e.g.,~drain and gate voltages, temperature, and process variation parameters) are directly mapped to their corresponding outputs (e.g.,~currents, charges, and capacitances). Prior studies have successfully employed \nn to model various types of transistors \cite{Tung2022, Guglani22, Tung2023, Zhang19, Eom2024, Thomann2024, Park24, Singhal25}, demonstrating the viability of this approach. Furthermore, \nn-based models have proven capable of meeting practical requirements, such as supporting variability analysis, \dtco \cite{Zhang19, Dai23}, switching dynamics \cite{Thomann2024}, and circuit simulation \cite{Tung2022, Tung2023}.

Most \nn-based approaches for transistor modeling employ \mlp architectures. The reliance on matrix multiplication in \mlp enables efficient training and fast inference due to its inherently simplistic computational nature. While the power of \mlp is sufficient to capture complex patterns and dependencies within transistors, it inherently lacks interpretability - a critical attribute for advancing scientific understanding. The ability to identify key parameters and features, analyze intricate data flows and transformations, and extract symbolic formulas would offer profound insights into transistor behavior, fostering further research and development. Consequently, a purely black-box approach may not be optimal for such applications.

Recently, Liu et al. introduced a novel architecture, \kan \cite{liu2024kan}, designed to enhance both interpretability and precision in physics-based function modeling \cite{Somvanshi24}. \kan draws inspiration from the Kolmogorov-Arnold representation theorem \cite{ka56, ka57} and innovatively employs learnable functions instead of traditional layer weights. These learned functions can be translated into symbolic forms, enabling the entire \kan architecture to be represented as a symbolic formula. This capability provides unparalleled interpretability, bridging the gap between precise modeling and meaningful insights.

\textbf{Our Novel Contributions:} This work pioneers the application of \kan and Fourier \kan architectures to transistor compact modeling, representing a significant advancement in this field. For the first time, we systematically evaluate their effectiveness in capturing critical transistor behaviors and figures of merit. We provide a comprehensive analysis of the strengths and limitations of \kan architectures, identifying unique challenges encountered during the modeling process and offering actionable insights for addressing them. Additionally, we benchmark \kan against the industry-standard compact model for FinFET technology as well as \mlp, emphasizing key performance metrics. Unlike prior works that often focus on current predictions, our evaluation rigorously assesses both current and charge predictions, expanding the scope of applicability in capturing the underlying physics. Finally, we demonstrate the remarkable capability of \kan to generate symbolic formulas from learned data patterns, showcasing its potential to deliver interpretable insights that drive both scientific discovery and practical advancements in transistor design.

\section{Our Proposed \kan Architecture for Transistor Compact Modeling}  \label{sec:arch}
Section \ref{sec:arch_mlp} briefly describes \mlp architecture for a better understanding of the concepts behind \kan and \fkan provided in Section \ref{sec:arch_kan} and Section \ref{sec:arch_fkan} respectively. Prior NN-based transistor compact modeling approaches and techniques are presented in Section \ref{sec:nn_transistor_modeling}.

\subsection{Multi-Layer Perceptron (\mlp)} 
\label{sec:arch_mlp}

\begin{figure}
\centering
\begin{subfigure}[b]{0.4\textwidth}
   \caption{\kan}
   \includegraphics[width=1\linewidth]{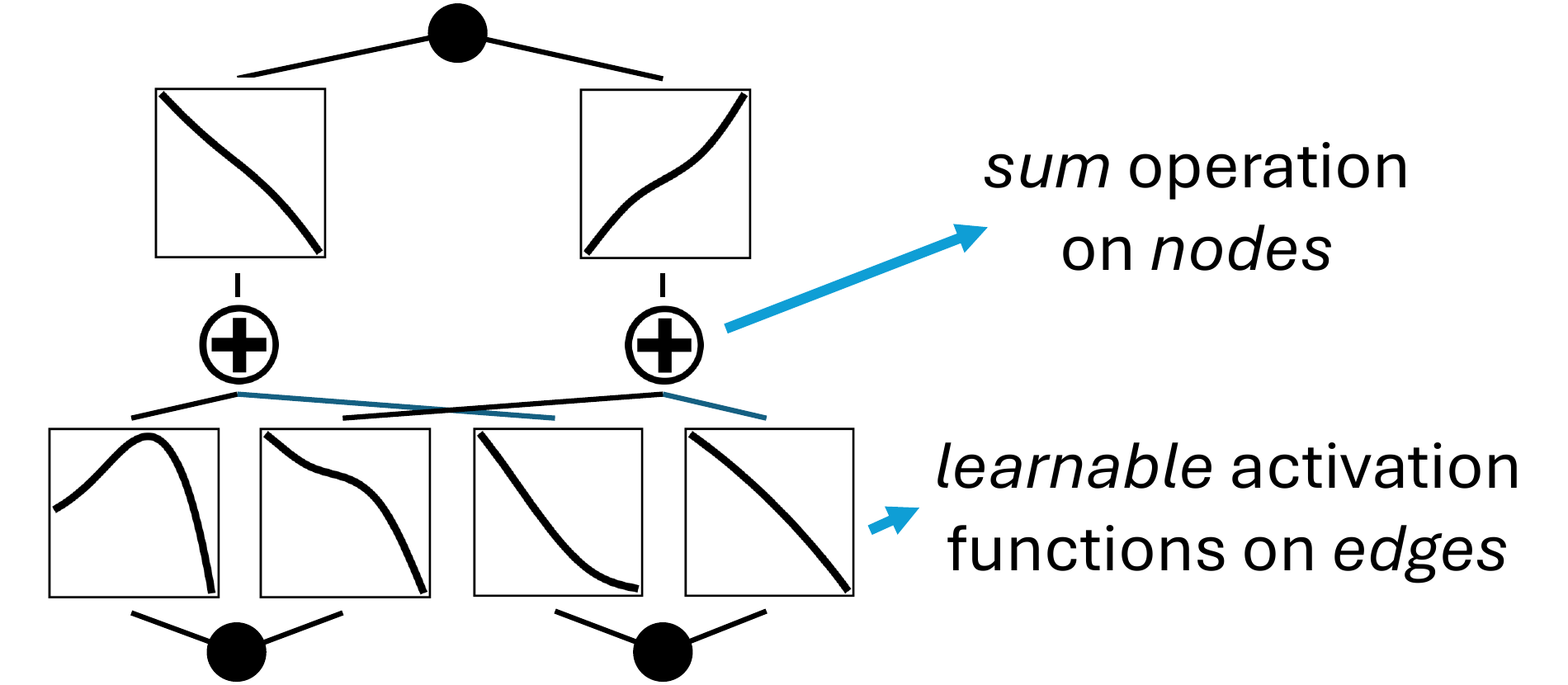}
   \label{fig:kan_arch} 
\end{subfigure}

\begin{subfigure}[b]{0.4\textwidth}
   \caption{\mlp}
   \includegraphics[width=1\linewidth]{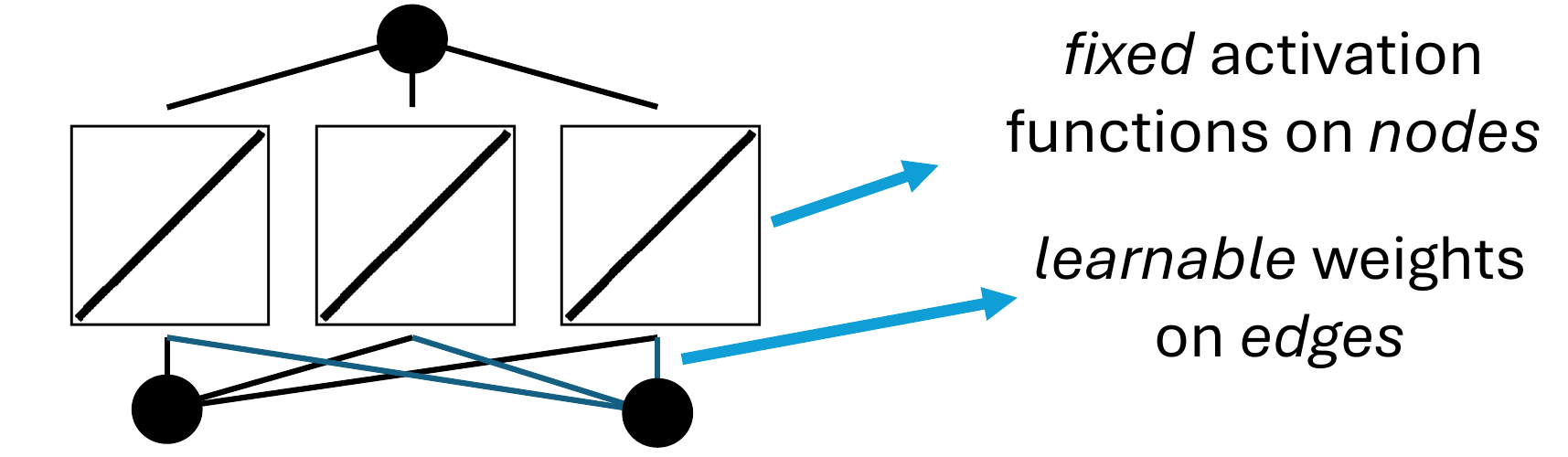}
   \label{fig:mlp_arch}
\end{subfigure}
\caption{\kan architecture (a), compared to \mlp architecture (b), employs learnable activation functions on edges instead of weights.}
\label{fig:kan_vs_mlp_arch}
\end{figure}


To understand the concept behind KANs, it is helpful to briefly revisit the functionality of \mlp. Essentially, an MLP is a class of feedforward NNs consisting of $N$ fully connected layers, which transform the input using a combination of linear and non-linear operations. Each layer $l_i$ is parameterized by a weight matrix $W_i$ and a bias vector $b_i$. These layers are usually followed by non-linear activation functions $\sigma_i$ (see Fig. \ref{fig:mlp_arch}). Activation functions $\sigma_i$ play a crucial role in MLPs, as they introduce non-linearity into the network, allowing it to approximate complex functions. Common choices for activation functions in NN-based transistor modeling include Rectified Linear Unit (ReLU) $ReLU(x) = \max(0, x)$, which helps mitigate the vanishing gradient problem, or Hyperbolic Tangent $tanh(x) = \frac{e^x - e^{-x}}{e^x + e^{-x}}$, which limits outputs to a range from -1 to 1. The computation performed by a single linear layer and activation function can be expressed as:
\begin{equation}\label{eqn:mlp_layer}
l_i(x) = \sigma_{i}(W_{i}\cdot x + b_i)
\end{equation}

If we assume a uniform activation function $\sigma$ across all layers and omit biases $b_i$ for simplicity, the forward pass of the MLP can be expressed as:
\begin{equation}\label{eqn:mlp_general_form1}
MLP(x) = W_N \cdot \sigma(W_{N-1}\cdot \sigma(...(W_2\cdot \sigma(W_1\cdot x))...))
\end{equation}
which can alternatively be written using function composition operators as:
\begin{equation}\label{eqn:mlp_general_form2}
MLP(x) = (W_N \circ \sigma \circ W_{N-1} \circ \sigma \circ...\circ W_2 \circ \sigma \circ W_1) \circ x
\end{equation}

While fixed activation functions proved to be sufficient for most of the tasks, learnable activation functions \cite{Apicella21, Tavakoli21} can potentially provide with adaptive non-linearity and higher expressive power. However, they greatly increase computational complexity and introduce additional optimization challenges.

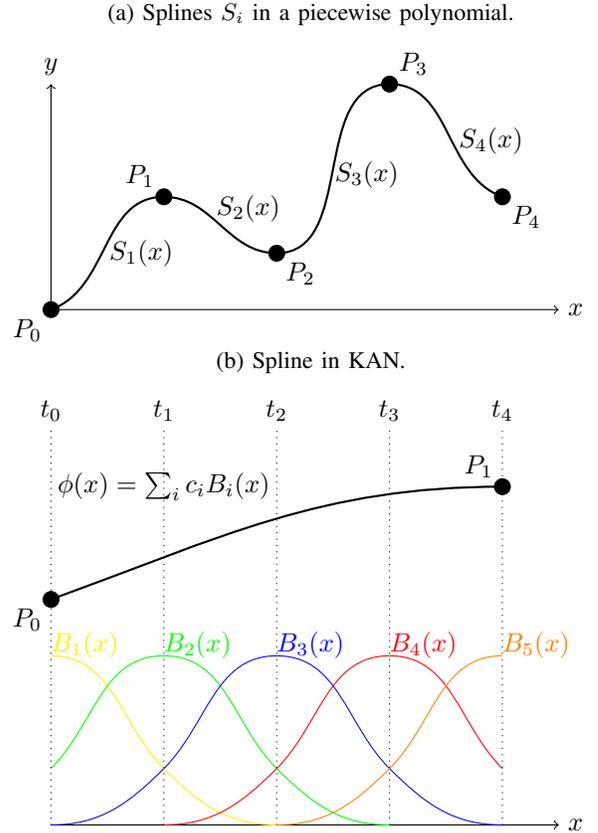
\begin{figure}
\centering
\begin{subfigure}[b]{0.45\textwidth}
\caption{Splines $S_i$ in a piecewise polynomial.}
    \begin{tikzpicture}[scale=1.5]
    \draw[->] (0, 1.0) -- (4.5, 1.0) node[right] {$x$};
    \draw[->] (0, 1.0) -- (0, 3.0) node[above] {$y$};
    \coordinate (A) at (0, 1);
    \coordinate (B) at (1, 2);
    \coordinate (C) at (2, 1.5);
    \coordinate (D) at (3, 3);
    \coordinate (E) at (4, 2);
    \draw[thick, black] (A) to[out=20, in=180] (B);
    \draw[thick, black] (B) to[out=0, in=180] (C);
    \draw[thick, black] (C) to[out=0, in=180] (D);
    \draw[thick, black] (D) to[out=0, in=160] (E);
    \filldraw[black] (A) circle (2pt) node[below left] {$P_0$};
    \filldraw[black] (B) circle (2pt) node[above left] {$P_1$};
    \filldraw[black] (C) circle (2pt) node[below right] {$P_2$};
    \filldraw[black] (D) circle (2pt) node[above right] {$P_3$};
    \filldraw[black] (E) circle (2pt) node[below right] {$P_4$};
    \node at (0.8, 1.5) {$S_1(x)$};
    \node at (1.75, 1.9) {$S_2(x)$};
    \node at (2.8, 2.2) {$S_3(x)$};
    \node at (3.9, 2.5) {$S_4(x)$};
    \end{tikzpicture}
    \label{fig:spline_math} 
\end{subfigure}

\begin{subfigure}[b]{0.45\textwidth}
\caption{Spline in \kan.}
    \begin{tikzpicture}[scale=1.5]
    \draw[->] (0, 0) -- (4.5, 0) node[right] {$x$};
    \draw[dotted] (0, 0) -- (0, 3.5) node[above] {$t_0$};
    \draw[dotted] (1, 0) -- (1, 3.5) node[above] {$t_1$};
    \draw[dotted] (2, 0) -- (2, 3.5) node[above] {$t_2$};
    \draw[dotted] (3, 0) -- (3, 3.5) node[above] {$t_3$};
    \draw[dotted] (4, 0) -- (4, 3.5) node[above] {$t_4$};
    \coordinate (A) at (0, 2);
    \coordinate (B) at (4, 3);
    \coordinate (S11) at (0, 1.5);
    \coordinate (S12) at (1, 0.5);
    \coordinate (S13) at (2, 0);

    \coordinate (S21) at (1, 1.5);
    \coordinate (S22) at (2, 0.5);
    \coordinate (S23) at (3, 0);
    \coordinate (S24) at (0, 0.5);

    \coordinate (S31) at (2, 1.5);
    \coordinate (S32) at (3, 0.5);
    \coordinate (S33) at (4, 0);
    \coordinate (S34) at (1, 0.5);
    \coordinate (S35) at (0, 0);

    \coordinate (S41) at (3, 1.5);
    \coordinate (S42) at (4, 0.5);
    \coordinate (S44) at (2, 0.5);
    \coordinate (S45) at (1, 0);

    \coordinate (S51) at (4, 1.5);
    \coordinate (S54) at (3, 0.5);
    \coordinate (S55) at (2, 0);
    
    \draw[thick, black] (A) to[out=20, in=180] (B);
    \draw[yellow] (S11) to[out=0, in=135] (S12);
    \draw[yellow] (S12) to[out=-45, in=180] (S13);

    \draw[green] (S24) to[out=45, in=180] (S21);
    \draw[green] (S21) to[out=0, in=135] (S22);
    \draw[green] (S22) to[out=-45, in=180] (S23);

    \draw[blue] (S35) to[out=0, in=-135] (S34);
    \draw[blue] (S34) to[out=45, in=180] (S31);
    \draw[blue] (S31) to[out=0, in=135] (S32);
    \draw[blue] (S32) to[out=-45, in=180] (S33);

    \draw[red] (S45) to[out=0, in=-135] (S44);
    \draw[red] (S44) to[out=45, in=180] (S41);
    \draw[red] (S41) to[out=0, in=135] (S42);

    \draw[orange] (S55) to[out=0, in=-135] (S54);
    \draw[orange] (S54) to[out=45, in=180] (S51);
    
    \filldraw[black] (A) circle (2pt) node[below left] {$P_0$};
    \filldraw[black] (B) circle (2pt) node[above left] {$P_1$};
    \node[yellow] at (0.3, 1.6) {$B_1(x)$};
    \node[green] at (1.3, 1.6) {$B_2(x)$};
    \node[blue] at (2.3, 1.6) {$B_3(x)$};
    \node[red] at (3.3, 1.6) {$B_4(x)$};
    \node[orange] at (4.3, 1.6) {$B_5(x)$};
    \node at (1.0, 3.0) {\textbf{$\phi (x)=\sum_i c_i B_i(x)$}};
    \end{tikzpicture}
    \label{fig:spline_kan} 
\end{subfigure}
\caption{Activation functions in \kan are parameterized $k$-order B-spline curves $\phi(x)$ with learnable coefficients $c_i$ of B-spline basis functions $B_i$. This implementation provides adaptive non-linearity and allows a better fit for complex physics-based functions.}
\label{fig:spline}
\end{figure}

\subsection{Kolmogorov-Arnold Network (\kan)} \label{sec:arch_kan}
In contrast to the \mlp, which has learnable weights $W_i$ on edges and fixed activation functions $\sigma$ on nodes (see Fig. \ref{fig:mlp_arch}), \kan employs learnable activation functions on edges with a simple sum operation on nodes (see Fig. \ref{fig:kan_arch}). \kan utilizes the \kart which states that every multivariate continuous function $f(x)$ in a bounded domain can be represented as a finite composition of continuous internal functions $\phi_{q,p}$ of a single variable $x_1,...,x_n$, external functions $\Phi_q$, and the addition operation:
\begin{equation}\label{eqn:kart}
    f(x)=f(x_1,...,x_n)=\sum_{q=0}^{2n}\Phi_q\left(\sum_{p=1}^n\phi_{q,p}(x_p)\right)
\end{equation}

Therefore, in order to approximate any function, parameters for functions $\phi$ and $\Phi$ have to be learned. However, these functions may not always be learnable in practice due to being non-smooth or fractal. Thus, in \kan they are represented as parameterized $k$-order B-spline curves $\phi(x)=\sum_i c_i B_i(x)$ with learnable coefficients $c_i$ of B-spline basis functions $B_i$ (See Fig. \ref{fig:spline_kan}), which has a similar concept with a standard piecewise polynomial shown in Fig. \ref{fig:spline_math}. The number of basis functions is $G+k$, where $G$ is the grid size. \kan layer $L$, represented by a set of edges with neighboring input and output nodes, consists of multiple learnable functions $\phi_{L,j,i}$. The number of $\phi_{L,j,i}$ is the number of input variables (nodes) $x_{L,i}$ multiplied by the number of neurons (edges) $n_L$ (see Fig. \ref{fig:kan_arch} with $n_0=2,n_1=2,n_2=1$). Thus, the output $y_{L,j}$ of a \kan layer $L$ is:
\begin{equation}\label{eqn:kan1}
    y_{L,j}=x_{L+1,j}=\sum_{i=1}^{n_{L}}\phi_{L,j,i}(x_{L,i}); \ \ j=1,...,n_{L+1}
\end{equation}
Note that the right side of Eq. \ref{eqn:kan1} is similar to the bracket term in \kart in Eq. \ref{eqn:kart}, which means that \kart is essentially a two-layer \kan with $n_0=n,n_1=2n+1,n_2=1$. Eq. \ref{eqn:kan1} rewritten in a matrix form yields:
\begin{equation}\label{eqn:kan2}
    x_{L+1} = \begin{pmatrix}
    \phi_{L,1,1} & \phi_{L,1,2} & ... & \phi_{L,1,n_L}\\
    \phi_{L,2,1} & \phi_{L,2,2} & ... & \phi_{L,2,n_L}\\
    ... & ... & ... & ...\\
    \phi_{L,n_{L+1},1} & \phi_{L,n_{L+1},2} & ... & \phi_{L,n_{L+1},n_L}\\
    \end{pmatrix} x_{L}
\end{equation}
\begin{equation}\label{eqn:kan3}
    x_{L+1} = \Phi_L x_{L}, \ \ x_{L} = \Phi_{L-1} x_{L-1},\ \ ...
\end{equation}
where $\Phi_L$ is a function matrix for layer $L$. Given the Eq. \ref{eqn:kan3}, a general \kan can be written with a composition operators as:
\begin{equation}\label{eqn:kan4}
    KAN(x) = (\Phi_{L-1} \circ \Phi_{L-2} \circ ... \circ \Phi_{1} \circ \Phi_{0}) \circ x
\end{equation}
which is very similar to Eq. \ref{eqn:mlp_general_form2}, meaning that \mlp treats linear and non-linear transformations separately as $W$ and $\sigma$ while \kan combines them under $\Phi$. To make \kan more optimizable, an additional basis function $b(x)$ is included in $\phi$:
\begin{equation}\label{eqn:kan5}
   \phi(x) = w_b b(x) + w_s spline(x)
\end{equation}
Trainable weights $w_b$ and $w_s$ allow better control over the influence of different parts of $\phi$. $b(x)$ can be chosen, and the original \kan implementation used Sigmoid Linear Unit (SiLU) $b(x)=\frac{x}{1+e^{-x}}$ \cite{Hendrycks2016}. As a result of having trainable B-spline coefficients and additional weights in $\phi$, \kan is generally more difficult to train than a \mlp, which employs only matrix multiplications and fixed non-linear activation functions. The original \kan architecture was later extended to MultKAN \cite{liu2024kan2}, which introduced multiplication operations on nodes. However, in this work, we will mainly focus on classic KANs.



\subsection{Fourier KAN (\fkan)} \label{sec:arch_fkan}
Since the essence of \kan is to replace any arbitrary complex function with multiple simple nonlinear functions, other representations were also explored in search of easier and potentially more efficient training. Xu et al. employed a Fourier transform \cite{Nussbaumer1982} and proposed a \fkan layer \cite{xu2024fourierkangcf} with the following replacement of $\phi$:
\begin{equation}\label{eqn:fkan1}
    \phi_F(x)=\sum_{i=1}^{d}\sum_{k=1}^G(cos(kx_i)\cdot a_{ik}+sin(kx_i)\cdot b_{ik})
\end{equation}
where $d$ is the number of neurons, $a_{ik}$ and $b_{ik}$ are trainable Fourier coefficients and $G$ is a grid size, which determines how many different sine and cosine terms are included in the Fourier transform corresponding to each input dimension. It simplifies the training procedure by replacing $k$-order splines and has a minor advantage in computational efficiency over regular \kan.

\subsection{NN-based Transistor Modeling} \label{sec:nn_transistor_modeling}
As was mentioned previously, most of NN-based transistor modeling approaches employ \mlp architectures due to their simplicity and fast inference. Nevertheless, many prior studies proved that the power of MLP is still sufficient to capture transistor behavior.

Tung et al. \cite{Tung2022} models $I-V$ and $C-V$ profiles of GAAFET from drain and gate voltages by designing a new loss function to obtain smooth higher-order derivatives with excellent accuracy. Additionally, a higher speed of NN-based approach over the standard BSIM model was demonstrated in Python. Later, this approach was extended with additional process variation parameters, i.e. gate length, fin height, equivalent oxide thickness, and work function difference \cite{Tung2023}. These parameters were added to the input to model FinFET charges, drain and gate currents with high precision. As before, NN-based models proved to be capable of supporting circuit simulation and demonstrated fast calculation speed in Python. Since adding more input variables naturally increases NN size due to the rising complexity of the task, Eom et al. proposed a flexible drain current modeling framework that employs multiple sets of small MLPs instead of one big NN \cite{Eom2024}. A small current prediction network (CPN), in conjunction with the bias modification network (BMN), used drain and gate voltages as inputs to predict the drain current of a-IGZO thin-film transistor, while multiple parameter generation networks were modifying CPN and BMN weights based on the selected device characteristics. Even though the number of process variation parameters went up to nine, the proposed NN framework was able to achieve $0.27\%$ relative error for drain current prediction. Further research was mainly focused on expanding NN-based frameworks to self-heating \cite{Tung24sh}, adding source and body voltages \cite{Singhal25}, integrating NN-based models into SPICE software by optimizing matrix multiplication in Verilog-A \cite{Tung24spice}, and modeling unsupported by BSIM transistors, e.g. a buried-channel-array transistor \cite{Park24}. Additionally, the possibility of combining an analytical low-fidelity device model with NN was also explored \cite{Sheelvardhan24} with NN used as an augmentation and correction tool to reduce the error of the analytical low-fidelity model. NN assistance and hybrid models for transistor compact modeling were also investigated by Kao et al. \cite{Kao22} and Wang et al. \cite{Wang24}.

While all previous approaches involved MLPs and array-like input data types, other network architectures and data structures were also employed in prior research. Yang et al. proposed graph-based compact modeling where physical components are conceptualized as graph nodes, and their respective interactions are depicted as graph edges. This ensures that these components and their correlations are maintained while allowing sufficient flexibility and efficiency in NN model development. Jang et al. \cite{Jang23} transforms the unstructured TCAD mesh into a graph with the further use of a graph neural network to estimate the electrostatic potential, electron/hole densities, and the \ids-\vgs curve of a MOSFET, effectively mimicking the TCAD simulation with a significant speed increase. Thomann et al. \cite{Thomann2024} analyzed switching dynamics of FeFET using convolutional NN for current and polarization map prediction. 

Despite their simplicity, MLPs still remain state-of-the-art architecture for NN-based transistor compact modeling. However, regular NNs inherently lack interpretability and act as a black box solver. It hinders the advancement in scientific understanding and identification of key features and parameters. Therefore, we explore a novel \kan architecture, specifically designed with high interpretability and precision for physics-based tasks in mind.

\section{Experimental Setup} 
\label{sec:exp}
This section details the dataset generation process, the train/test data split, and the network parameters used in our experiments. Section \ref{sec:exp_nparams} provides an overview of the networks and their configurations, while Section \ref{sec:exp_dconversion} describes the conversion functions essential for current and charge fitting.

\subsection{Dataset Generation} 
\label{sec:exp_dataset}
To evaluate the performance of KANs and MLPs, we utilize a $7$~nm FinFET technology \cite{Clark16}. For a single device, we model the drain current \ids, drain charge \qd, and source charge \qs characteristics by varying the drain voltage \vds and gate voltage \vgs from 0 to 0.82 V with a step size of 5 mV. The data is generated using a commercial SPICE tool (HSPICE) \cite{hspice_cit} and the industry-standard compact model for FinFET technology, which is internally integrated (MOSFET Level 72).
Obtained data is divided into four datasets to assess the networks' generalization capabilities. The first dataset includes all $27,225$ data points in the training set, leaving the test set empty. The remaining datasets are constructed by varying the step sizes of \vds and \vgs in their respective training sets, with the excluded points allocated to the test sets. An overview of all datasets is provided in Table \ref{table:data_overview}.

\begin{table}
\centering
\caption{Datasets generated with industry-standard compact model.}
\label{table:data_overview}
\resizebox{\columnwidth}{!}{%
\begin{tabular}{ccccc}
\toprule
No. & Voltage step, $mV$   & Train points & Test points & Train/(Train+Test), $\%$ \\ \midrule
1 & 5 & 27225 & - & 100.0 \\ 
2 & 10 & 6889 & 20336 & 25.3 \\ 
3 & 20 & 1764 & 25461 & 6.48 \\ 
4 & 50 & 289 & 26936 & 1.06 \\ 
\end{tabular}}
\end{table}

\begin{table*}
\centering
\caption{Network overview.}
\label{table:net_overview}
\resizebox{1.8\columnwidth}{!}{%
\begin{tabular}{ccccc}
\toprule
 Network   & Description & Num. parameters & Inference time, ms \\ \midrule
MLP1  & $IN \rightarrow HL\rightarrow OUT$,\ \ $h$=16 & 337 & 0.04 \\ 
MLP2  & $IN \rightarrow HL\rightarrow HL\rightarrow OUT$,\ \ $h$=16 & 609 & 0.05 \\ 
KAN1  & $I_D$:[2,3,1,1],\ \ $Q$:[2,3,1],\ \ $k$=3,\ \ $G$=16 & $I_D$:338,\ \ $Q$:295 & 3 \\ 
KAN2  & $I_D$:[2,3,3,1,1],\ \ $Q$:[2,3,3,1],\ \ $k$=3,\ \ $G$=16 & $I_D$:587,\ \ $Q$:544 & 4.8 \\ 
FKAN1  & $IN\rightarrow OUT$, \ \ $h$=8, \ \ $G$=8 & 393 & 0.3 \\ 
FKAN2  & $IN\rightarrow HL\rightarrow OUT$, \ \ $h$=8, \ \ $G$=8, \ \ $G_{HL}$=2 & 657 & 0.5 \\  \bottomrule
\end{tabular}}
\end{table*}

\subsection{Neural Network Architecture Overview} 
\label{sec:exp_nparams}
To analyze how \textit{scaling} the number of network parameters affects prediction error, we employ multiple models of each network type in our experiments. Similar to other works we use relatively small MLPs to maintain the idea of fast and compact transistor modeling. \textbf{(1)} The MLP architectures feature a hidden dimension of $h=16$ neurons. The smaller MLP has one hidden layer, while the larger one includes two hidden layers. The $Tahn$ activation function is used to constrain the output values of linear layers.
\textbf{(2)} For the small \ids \kan model, we selected a configuration with $n_0=2, n_1=3, n_2=1, n_3=1$. The larger \ids \kan model includes an additional hidden layer, resulting in $n_0=2, n_1=3, n_2=3, n_3=1, n_4=1$. For charge modeling, the last layer with one edge was removed. All KAN architectures use a spline order of $k=3$ and a grid size of $G=16$. \textbf{(3)} The parameter count in \fkan scales rapidly due to each spline requiring two coefficients, $a_{ik}$ and $b_{ik}$, for every function $\phi$ in each neuron (inner and outer sums in Eq. \ref{eqn:fkan1}). To ensure fair comparisons, the smaller \fkan model comprises only input and output layers with a hidden dimension of $h=8$ and grid size $G=8$. The larger \fkan model adds a hidden layer, with the grid size reduced to $G=2$ to manage the parameter count.

Table \ref{table:net_overview} provides a comprehensive summary of all network configurations, including their number of learnable parameters and inference times. The dimensions of the \kan layers, $n_i$, are represented as arrays, while $IN$, $OUT$, and $HL$ denote input, output, and hidden layers, respectively.

\subsection{Data Conversion} 
\label{sec:exp_dconversion}

Following prior works \cite{Tung2022,Tung2023,Eom2024}, we use conversion functions to constrain and adapt the output range of networks. The conversion functions for \ids, \qd, and \qs, corresponding to the network outputs $y_I$ and $y_Q$, are defined as:

\begin{equation}\label{eqn:cf1}
    I_D = e^{y_I}
\end{equation}
\begin{equation}\label{eqn:cf2}
    Q_{D,S} = y_Q \cdot 10^{-18}
\end{equation}

The conversion function for \ids enables the networks to operate on a logarithmic scale. As previously noted, the negative \ids region is not considered in this study. Therefore, we omit the \vds term in Eq. \ref{eqn:cf1}, which was used in earlier works \cite{Tung2022,Tung2023,Eom2024} to determine the sign of \ids. For charge modeling, while charge values can be negative, their signs cannot be easily determined. To address this, all charge values are scaled into a range that ensures efficient network operation.
Unlike current modeling, logarithmic scaling is unnecessary for charge modeling since charges do not exhibit the same variability as \ids within the given voltage range. Moreover, loss functions designed for logarithmic-scale training are incompatible with signed data, as the logarithm of a negative number is undefined. This necessitates the straightforward scaling approach used for charges.


\section{Evaluation, Experimental Results, and Comparisons}
\label{sec:results}

This section presents the evaluation methodology, experimental results, and comparisons across the tested architectures. Section \ref{sec:training} details the training procedures for all networks. Performance metrics, including errors and derivatives, are analyzed in Sections \ref{sec:perror} and \ref{sec:derivs}, respectively. The symbolic regression capabilities of \kan are discussed in Section \ref{sec:sreg}, and in Section \ref{sec:isreg}, we present an algorithm for improving \kan symbolic regression capabilities. Finally, circuit simulation using FKAN transistor model is performed in Section \ref{sec:hspice}.

\subsection{Neural Network Training} 
\label{sec:training}

All networks are trained using the PyTorch framework \cite{pytorch}. For each network type listed in Table \ref{table:net_overview} and each dataset described in Table \ref{table:data_overview}, we train 20 models by varying the random seed. This approach allows us to investigate the \textit{convergence} properties of all three architectures and assess how reliably each architecture achieves optimal accuracy under identical training conditions. Furthermore, we train networks with varying trainable parameter numbers to explore the correlation between models' performance and their respective sizes.
The batch size is set to the full length of the training dataset in all cases, enhancing training stability. Other training parameters, such as learning rate and optimizer settings, differ depending on the network type to ensure optimal performance for each architecture.

All networks are trained with the following loss functions for currents and charges:
\begin{equation}\label{eqn:loss1}
\begin{split}
    L_I = a\cdot Er(I_D) + Er(log(I_D)) + Er(g_m) + \\ + Er(g_{DS}) + Er(g'_m) + Er(g'_{DS})
\end{split}
\end{equation}
\begin{equation}\label{eqn:loss2}
    L_Q = Er(Q) + Er(\partial Q/\partial V_G) + Er(\partial Q/\partial V_D)
\end{equation}
\begin{equation}\label{eqn:mse}
    Er(Y)=\frac{1}{p}\sum_{i=1}^p(y_i-\hat{y_i})^2
\end{equation}

Similar to previous approaches \cite{Tung2022,Tung2023,Eom2024}, we incorporate \textit{gradient information} to optimize model fitting. The parameter $a$ enhances the influence of \ids values in the saturation region and is set to $100$ across all experiments. Since all MLPs are trained in a single iteration, we employ the Adam optimizer with a weight decay of $wd=10^{-5}$ to achieve optimal fitting. The initial learning rates are set to $lr_I=0.005$ for \ids and $lr_Q=0.01$ for charges. The learning rate is gradually reduced during training when the loss function shows no further improvement. Once the learning rate drops below $10^{-5}$, weight updates become negligible, and training is terminated. In practice, all MLPs converged after approximately $40,000$ epochs.

For \kan training, we utilize the PyKAN framework \cite{liu2024kan} with the LBFGS optimizer. The learning rates are set to $lr_I=0.1$ for \ids and $lr_Q=1.0$ for charges. While the final grid size is $G=16$, we employ a grid refinement process to improve prediction results. Grid refinement involves iteratively extending the grid, where a coarser grid is used to initialize and refine a more precise one. In our experiments, the refinement starts with $G=2$ and progresses through $G\in [2,4,8,12,16]$ until reaching $G=16$. Experimentation determined $G=16$ as the optimal size, as larger grids increased the loss. Each grid step involves 300 epochs of training, totaling 1500 epochs for the entire refinement process.
For \fkan training, we adopt a method similar to \mlp training employing the Adam optimizer, but the initial learning rate is lowered to $lr=0.002$. \fkan models are trained for 60,000 epochs, with the learning rate reduced by a factor of 0.85 every 2000 epochs. This approach results in a final learning rate of $lr=1.53\cdot 10^{-5}$, following the same scheduling principle as \mlp.
A direct comparison of training times across the networks is not feasible due to differences in logging mechanisms and frameworks. However, for reference, we provide inference times for all models in Table \ref{table:net_overview}. All times are measured in Python on an Nvidia RTX 3070 GPU with a batch of 27,225 data points. It is worth noting that PyKAN is not as optimized as the PyTorch implementation of \mlp at the time of writing. Nonetheless, the slower inference speed of \kan and \fkan is an inherent limitation, as both architectures rely on complex spline calculations within their neurons, unlike the simpler and highly optimized matrix multiplications used in \mlp. While KANs were trained in the PyKAN framework, FKANs were trained in a custom framework written in PyTorch. Thus, their inference speed is much higher.

\begin{figure*}[!ht]
\centering
\begin{subfigure}[b]{0.33\textwidth} 
    \caption{Drain current \ids, small NNs.} 
    \includegraphics[width=\linewidth]{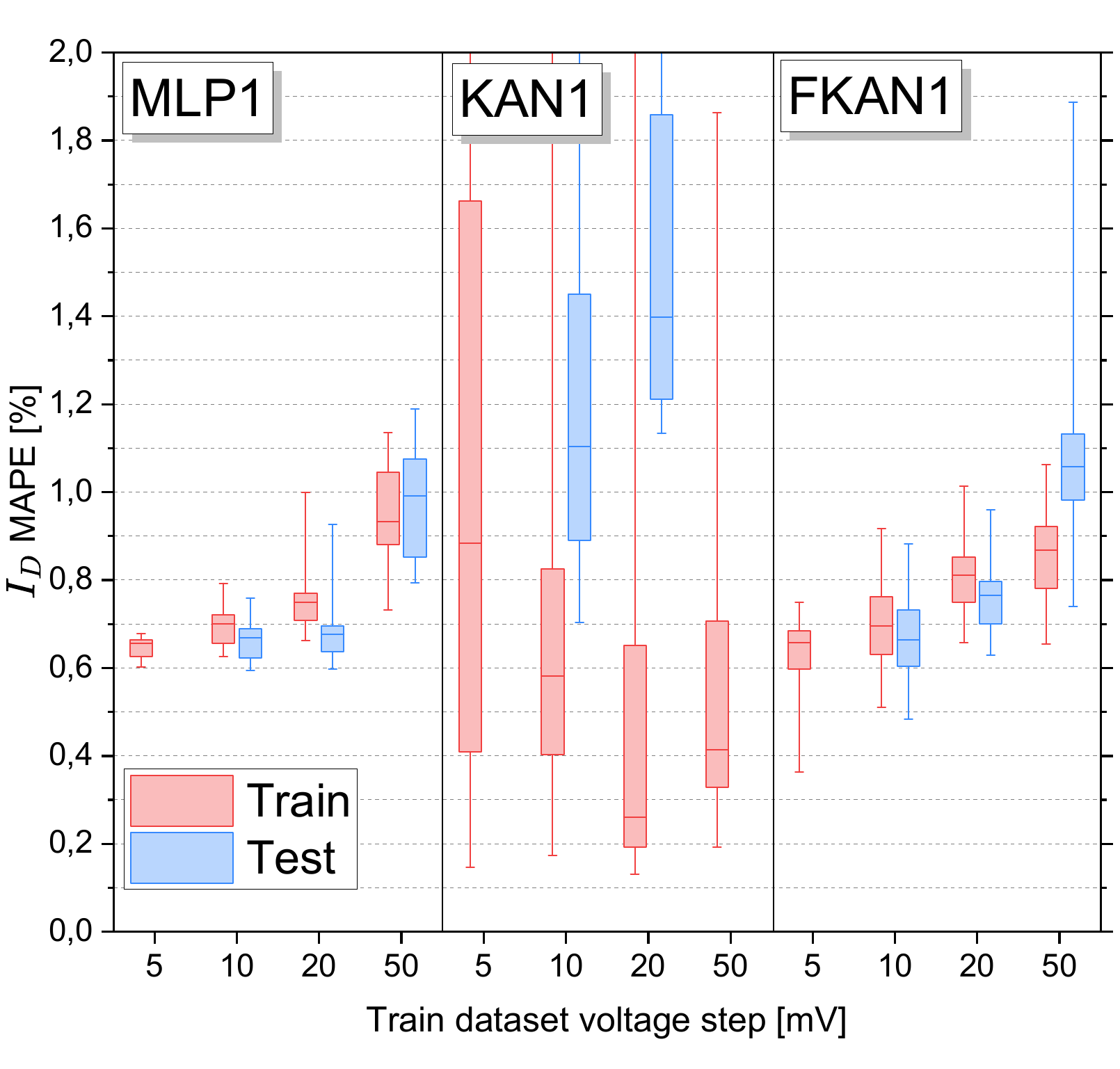}
    \label{fig:mape_plots_id}
\end{subfigure}\hspace{0.01em}%
\begin{subfigure}[b]{0.33\textwidth} 
    \centering
    \caption{Drain charge \qd, small NNs.} 
    \includegraphics[width=\linewidth]{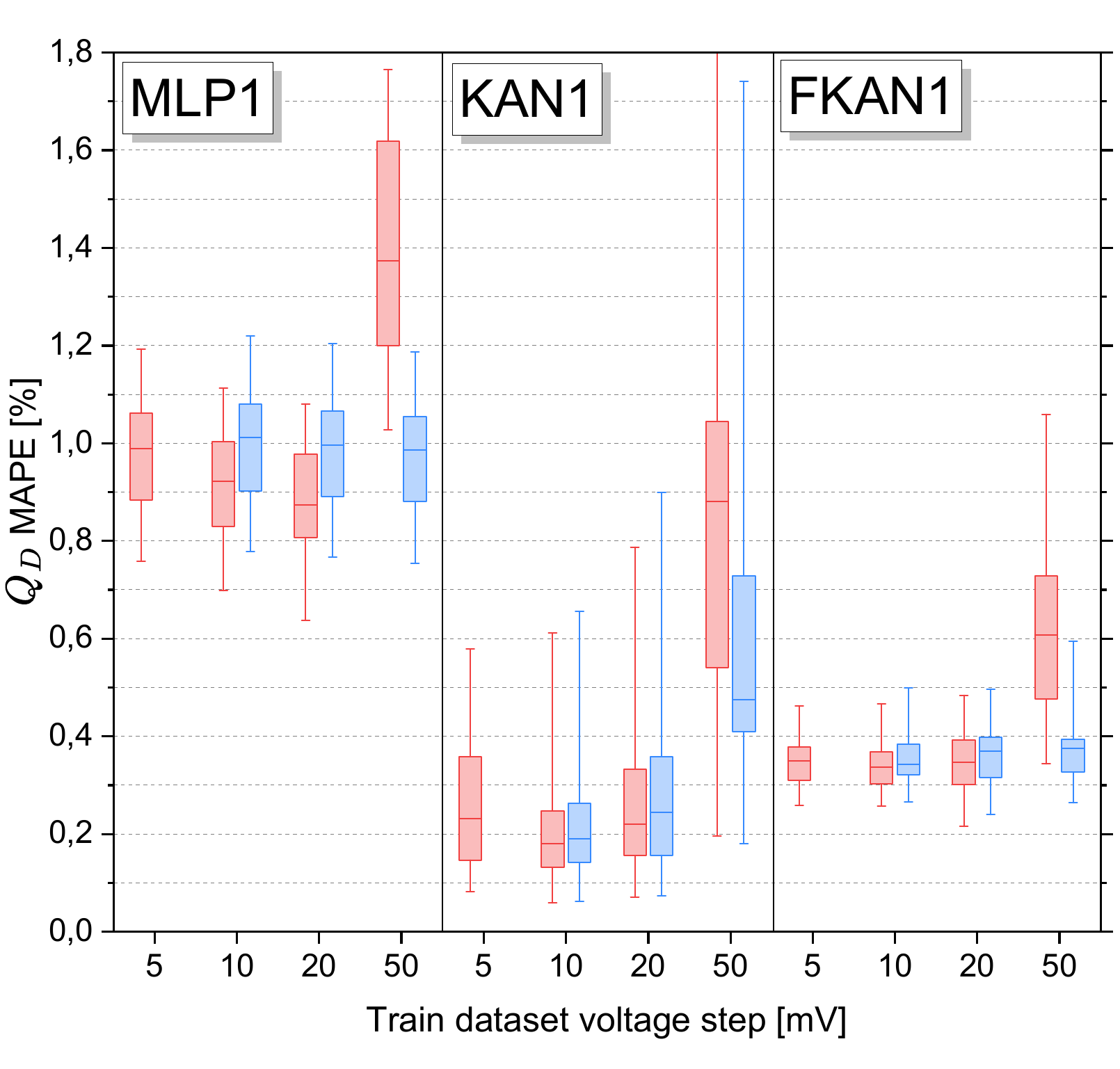}
    \label{fig:mape_plots_qd}
\end{subfigure}\hspace{0.01em}%
\begin{subfigure}[b]{0.33\textwidth} 
    \centering
    \caption{Source charge \qs, small NNs.} 
    \includegraphics[width=\linewidth]{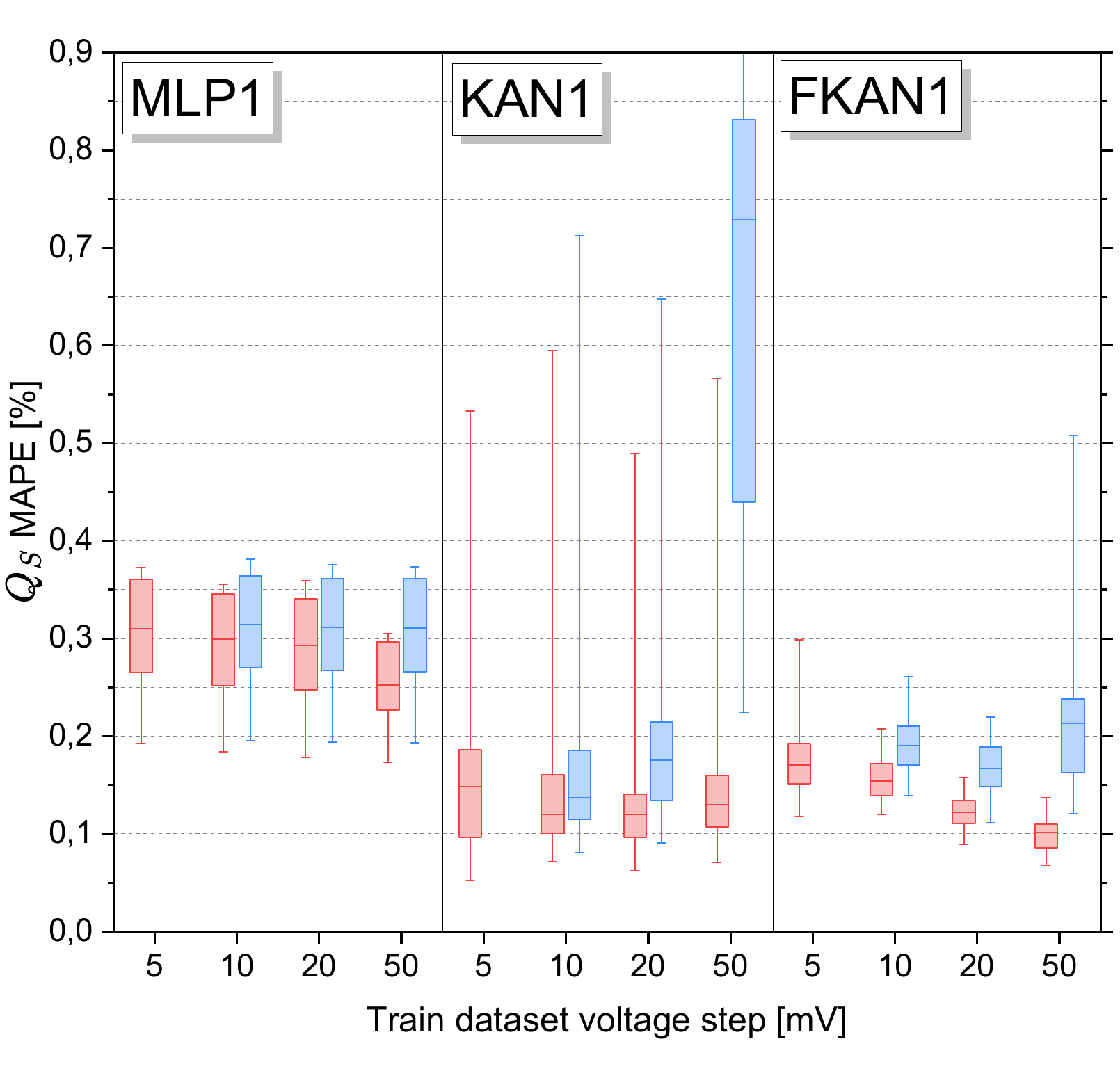}
    \label{fig:mape_plots_qs}
\end{subfigure}

\begin{subfigure}[b]{0.33\textwidth} 
    \caption{Drain current \ids, big NNs.} 
    \includegraphics[width=\linewidth]{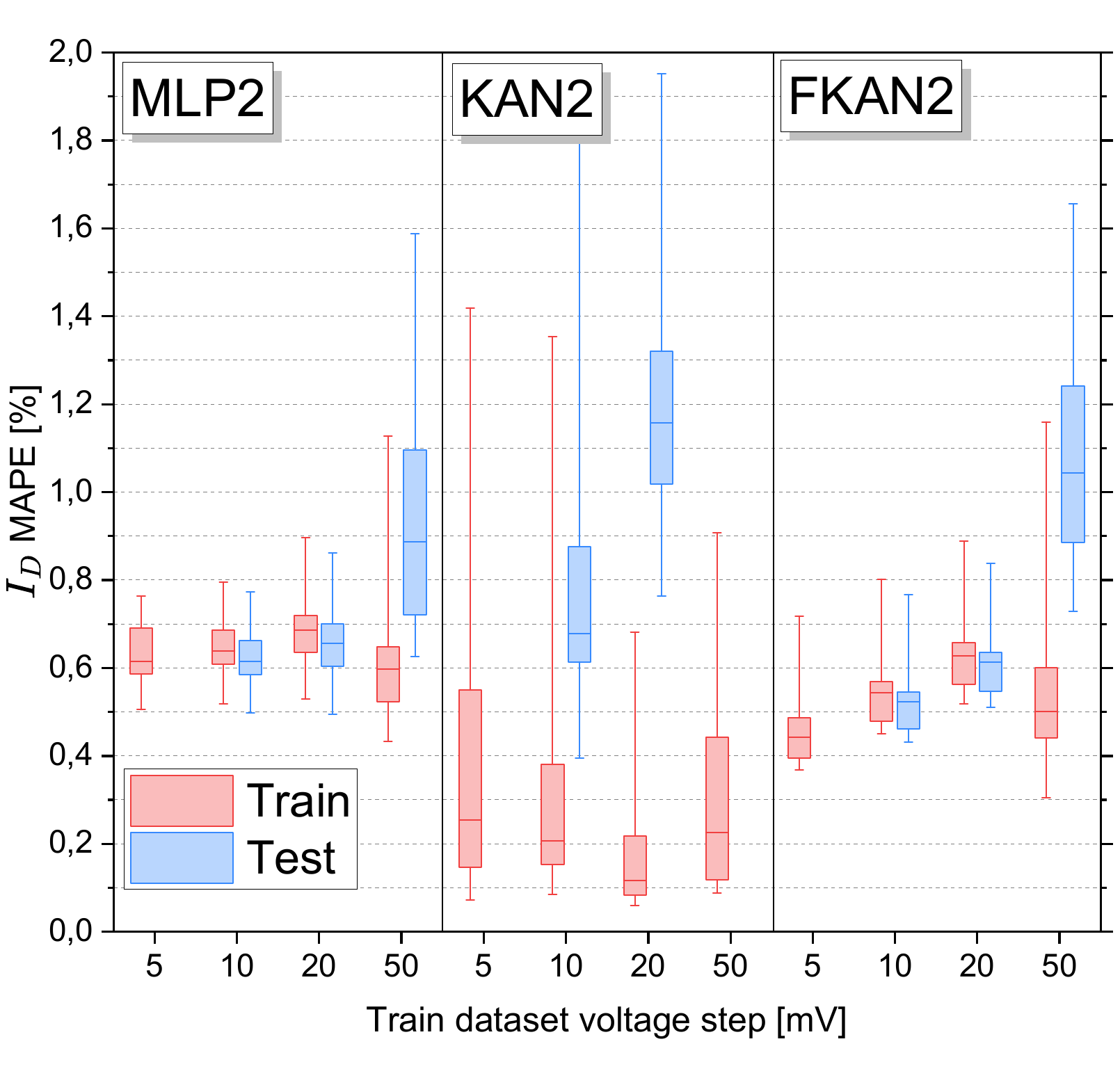}
    \label{fig:mape_plots_2h_id}
\end{subfigure}\hspace{0.01em}%
\begin{subfigure}[b]{0.33\textwidth} 
    \centering
    \caption{Drain charge \qd, big NNs.} 
    \includegraphics[width=\linewidth]{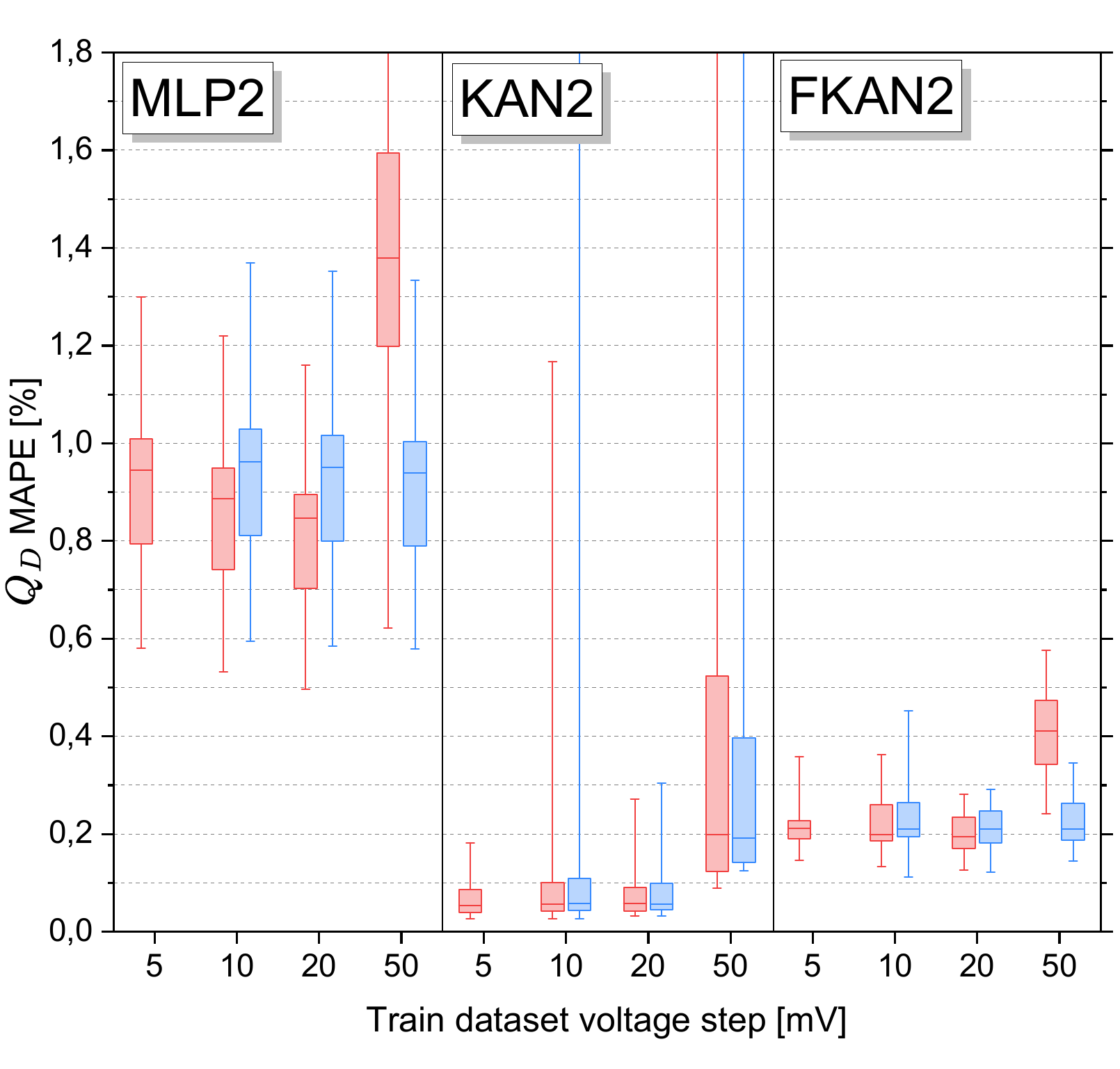}
    \label{fig:mape_plots_2h_qd}
\end{subfigure}\hspace{0.01em}%
\begin{subfigure}[b]{0.33\textwidth} 
    \centering
    \caption{Source charge \qs, big NNs.} 
    \includegraphics[width=\linewidth]{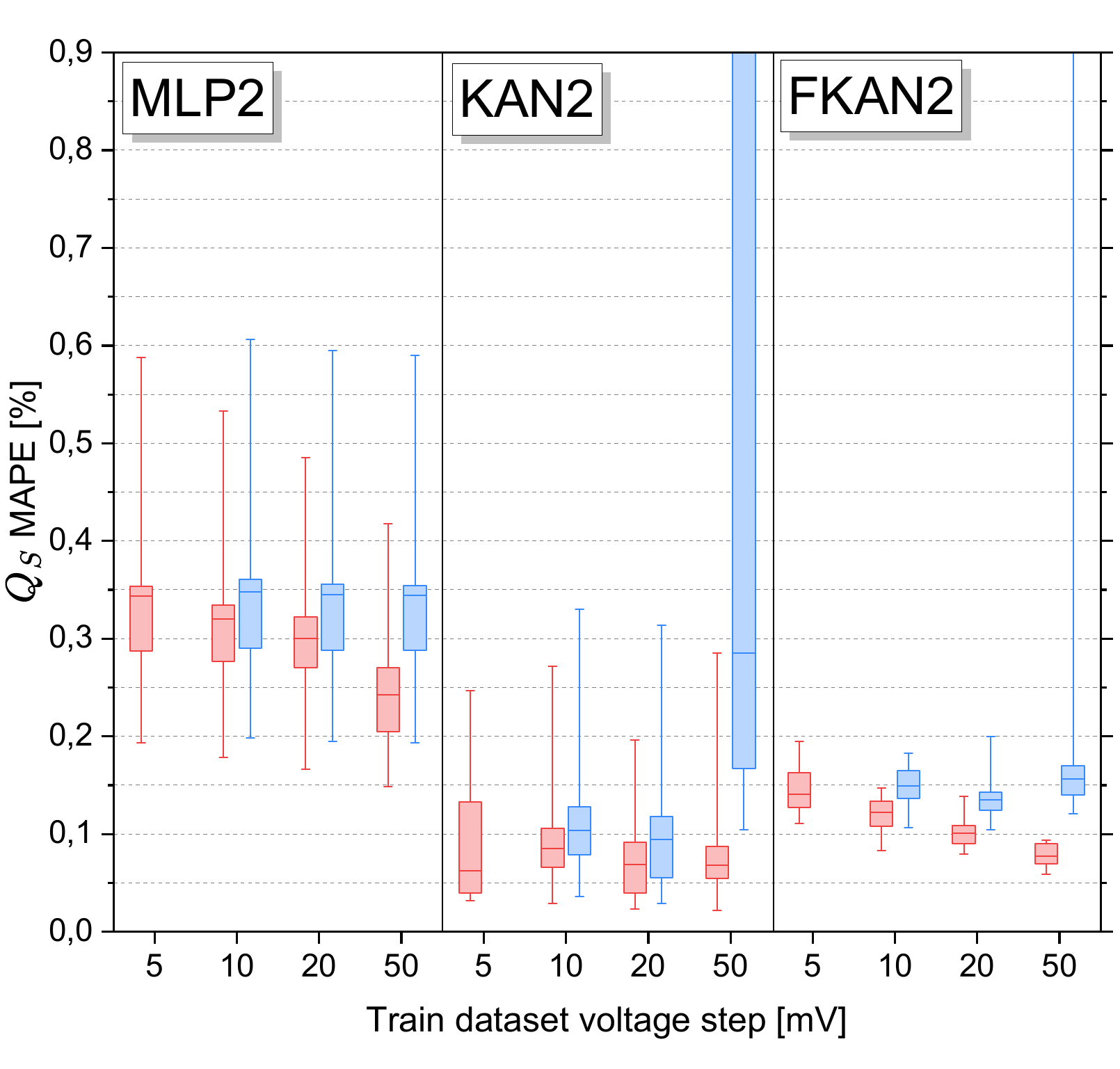}
    \label{fig:mape_plots_2h_qs}
\end{subfigure}
    \caption{Mean absolute percentage train (red) and test (blue) error (MAPE), calculated against industry-standard compact model. The comparison on various datasets from Table \ref{table:data_overview} for smaller (top) and bigger (bottom) networks shows KANs' remarkable fitting accuracy over MLPs and FKANs, while highlighting poor \ids generalization and lower consistency. Network names are architectures from Table \ref{table:net_overview}. Boxes correspond to 25,75 percentiles and represent consistency. The median value is illustrated with a middle line within each box. Whiskers show the minimum and the maximum error achieved by networks. Test MAPE for \ids and $50mV$ step in KAN1 and KAN2 cases exceeds plot limits and is not shown for better general visibility.}
\label{fig:mape_plots}
\end{figure*}

\subsection{Prediction Error Analysis} 
\label{sec:perror}

To evaluate a prediction accuracy on such a wide range of data as transistor currents and charges, we employ a ratio-based \mape for a predicted value $\hat{y}$ and a true value $y$, which represents simulation data of the industry-standard compact model in SPICE:
\begin{equation}\label{eqn:mape}
    \text{\mape}=\frac{\sum_{i=1}^{n} |y_i-\hat{y}_i|}{\sum_{i=1}^{n}|y_i|}
\end{equation}

Since charge prediction is not on a logarithmic scale (see Eq. \ref{eqn:cf2}) and achieving high precision for very low charge values is generally less critical, we exclude all charge values below $0.01\cdot 10^{-18}=10^{-20}F$ from the evaluation. This adjustment ensures that \mape remains consistent. For instance, $\hat{y}=0.002$ and $y=0.001$ in Eq. \ref{eqn:mape} yield a $100\%$ error, which could disproportionately skew the average \mape. 
The \mape results for the smaller architectures, MLP1, KAN1, and FKAN1, are illustrated in Fig. \ref{fig:mape_plots_id}-\ref{fig:mape_plots_qs}. Since the training error is influenced by the number of training points, we report the best training error based on the dataset with a $5mV$ voltage step size (see dataset No.~1 in Table \ref{table:data_overview}).

\textbf{For the drain current} \ids in Fig. \ref{fig:mape_plots_id}, MLP1 demonstrates the most consistent result, reaching $0.6\%$ train and test errors at best. The range of the 25,75-percentile box determines consistency, with the lower range corresponding to more consistent results and better convergence across 20 networks with different seeds. KAN1 reaches the lowest train error of $0.14\%$, but struggles to \textit{generalize} to the test data, only achieving $0.7\%$ test error. Additionally, the consistency for KAN1 is the worst among all other network types. By reaching $0.36\%$ train and $0.48\%$ test error, FKAN1 is a compromise between MLP1 and KAN1. The consistency and generalization capabilities of FKAN1 are lower than those of MLP1 but significantly better than those of KAN1, while the error value stays between.

\textbf{Charge modeling} \mape in Fig \ref{fig:mape_plots_qd} and Fig. \ref{fig:mape_plots_qs} demonstrates a similar tendency as for currents. Nevertheless, multiple significant improvements can be observed for KAN1 and FKAN1. Similar to \ids, MLP1 still shows good generalization and consistency but has the highest train and test errors among the three architectures - $0.76/0.7\%$ and $0.19/0.19\%$ for \qd and \qs correspondingly. In contrast to the previous case, KAN1 has a much better generalization, which allows to achieve the lowest train and test errors among all networks - $0.08/0.06\%$ for \qd and $0.05/0.08\%$ for \qs. FKAN1 with $0.25/0.23\%$ and $0.12/0.11\%$ \mape for \qd and \qs, respectively, remains in the middle but has the best consistency. 

\textbf{Current comparisons:} Figs~\ref{fig:mape_plots_2h_id}-\ref{fig:mape_plots_2h_qs} shows results for the bigger networks MLP2, KAN2, and FKAN2. While all tendencies observed in smaller networks persist, the idea behind training bigger models is to explore the scaling capabilities of various architectures. Thus, when writing the error values, we will also add error reduction over the respective error values from smaller networks. In the drain current \ids modeling (see Fig. \ref{fig:mape_plots_2h_id}), MLP2 achieves $0.5(-0.1)\%$ train and test errors, but consistency and generalization capabilities are slightly reduced compared to the smaller MLP1. KAN2 improves over the smaller KAN1 with $0.09(-0.05)\%$ train and $0.4(-0.3)\%$ test errors remaining the best model for a train data fit. Even though KAN2's consistency got better, poor generalization for \ids modeling still persists. As before, FKAN2 is in between two previous networks reaching $0.37(+0.01)\%$ train and $0.43(-0.05)\%$ test error. Thus, both MLP2 and KAN2 improve their accuracy over the smaller \ids models, while FKAN2's best performance remains quite similar. However, FKAN2's consistency shifted towards a lower error.

\textbf{Charge comparisons:} In contrast to the current, charge modeling in Fig. \ref{fig:mape_plots_2h_qd} and Fig. \ref{fig:mape_plots_2h_qs} demonstrates slightly different outcomes. MLP2 still shows a significant improvement for \qd with a $0.58 \mathclose{/}\mathopen{0.59} (-0.18 \mathclose{/}\mathopen{-0.11})\%$ train/test error but gets almost no benefits from the increased number of parameters for \qs reaching only $0.19\mathclose{/}\mathopen{0.20}(-0.0\mathclose{/}\mathopen{+0.01})\%$ errors. KAN2, on the other hand, significantly improves convergence and reduces the error down to $0.03\mathclose{/}\mathopen{0.03}(-0.05\mathclose{/}\mathopen{-0.03})\%$ for \qd and $0.03\mathclose{/}\mathopen{0.03}(-0.02\mathclose{/}\mathopen{-0.05})\%$ for \qs. Unlike \ids, FKAN2 charge modeling receives noticeable benefits from a larger model size, resulting in $0.14\mathclose{/}\mathopen{0.11}(-0.11\mathclose{/}\mathopen{-0.12})\%$ error for \qd and $0.11\mathclose{/}\mathopen{0.10}(-0.01\mathclose{/}\mathopen{-0.01})\%$ for \qs. It is worth mentioning that network \mape is generally harder to reduce the lower it gets, and the value reaches a plateau eventually. Thus, it is difficult to conclude which architecture benefits from the increased model size the most since a higher number of parameters gives advantages and disadvantages to all networks. However, we will summarize the outcomes for all architectures further below.

\begin{figure*}[!ht]
\centering
\begin{subfigure}[b]{0.33\textwidth} 
    \caption{Transconductance $g_m$ for MLP2.} 
    \includegraphics[width=\linewidth]{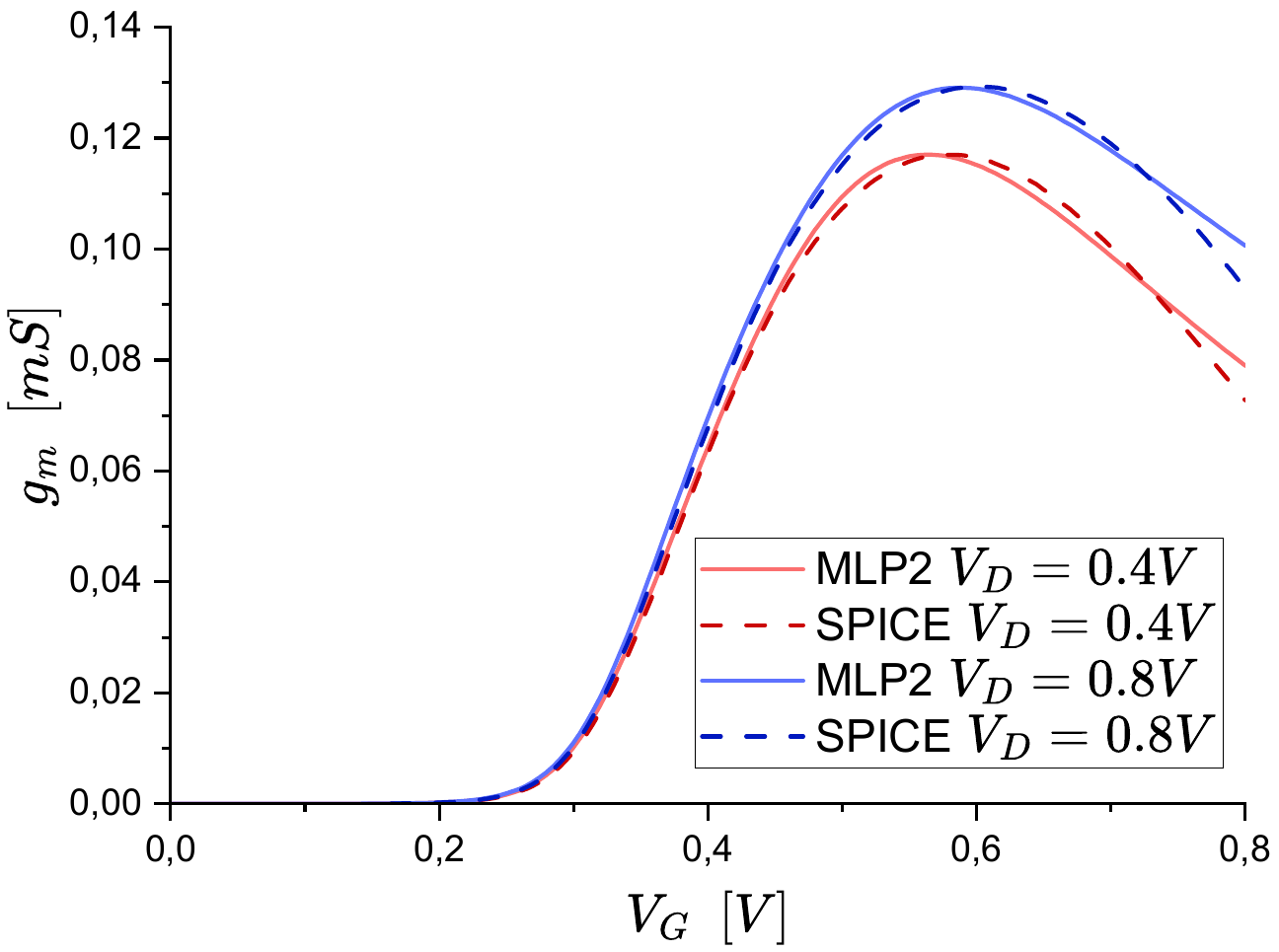}
    \label{fig:derivative_plots_d1_mlp}
\end{subfigure}\hspace{0.01em}%
\begin{subfigure}[b]{0.33\textwidth} 
    \centering
    \caption{Transconductance $g_m$ for KAN2.} 
    \includegraphics[width=\linewidth]{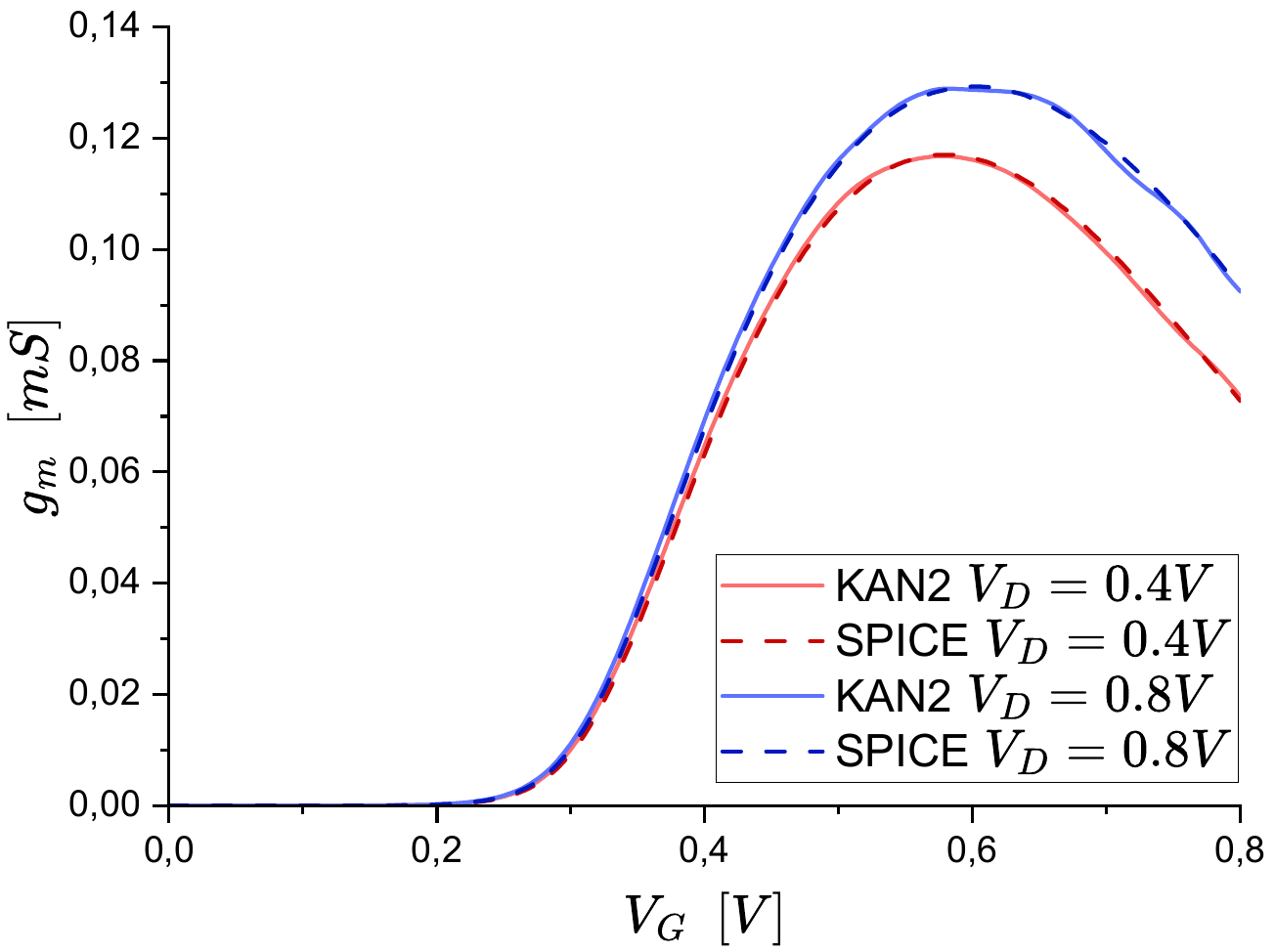}
    \label{fig:derivative_plots_d1_kan}
\end{subfigure}\hspace{0.01em}%
\begin{subfigure}[b]{0.33\textwidth} 
    \centering
    \caption{Transconductance $g_m$ for FKAN2.} 
    \includegraphics[width=\linewidth]{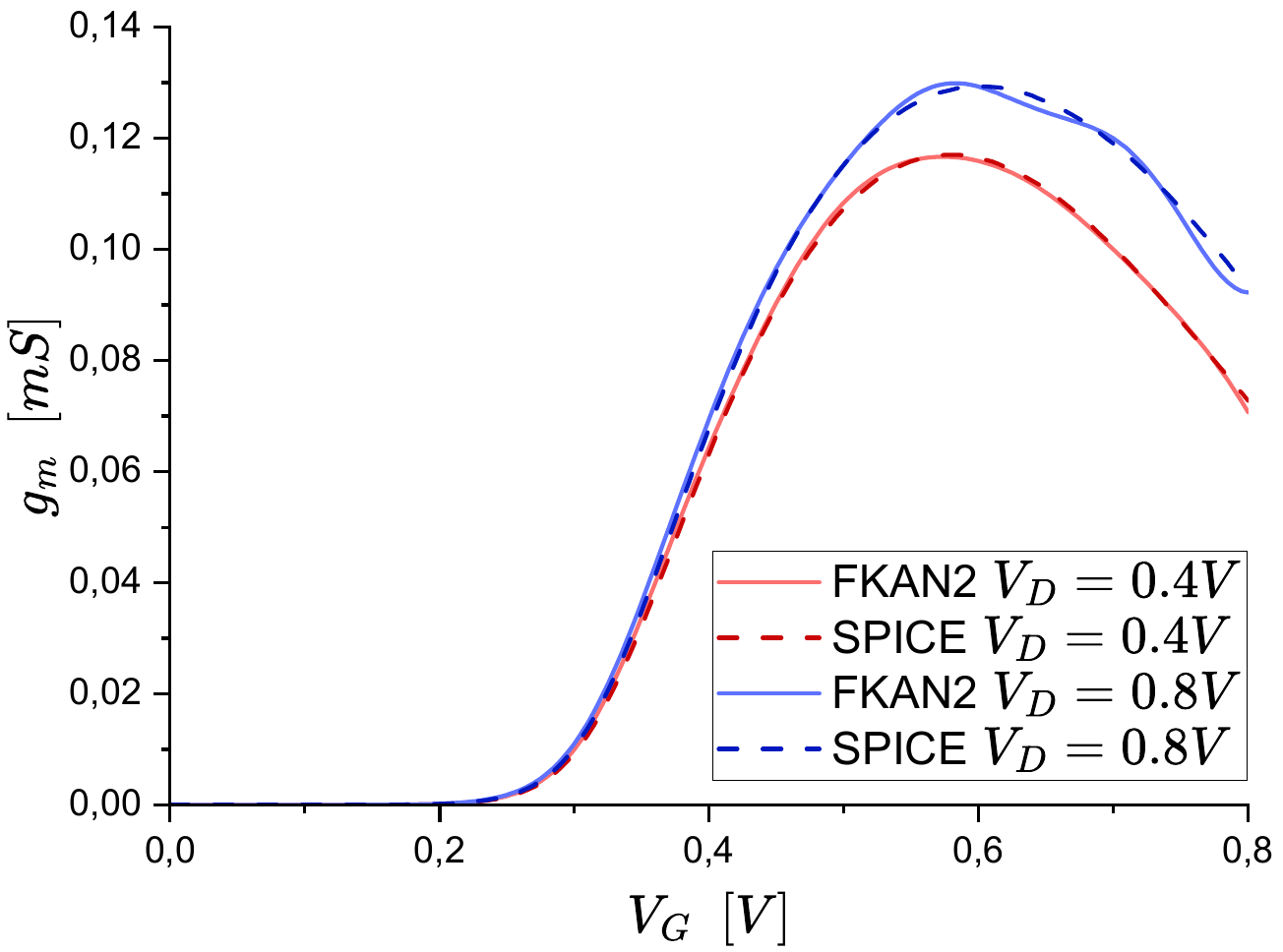}
    \label{fig:derivative_plots_d1_fkan}
\end{subfigure}

\begin{subfigure}[b]{0.33\textwidth} 
    \caption{$\partial g_m/ \partial V_g$ for MLP2.} 
    \includegraphics[width=\linewidth]{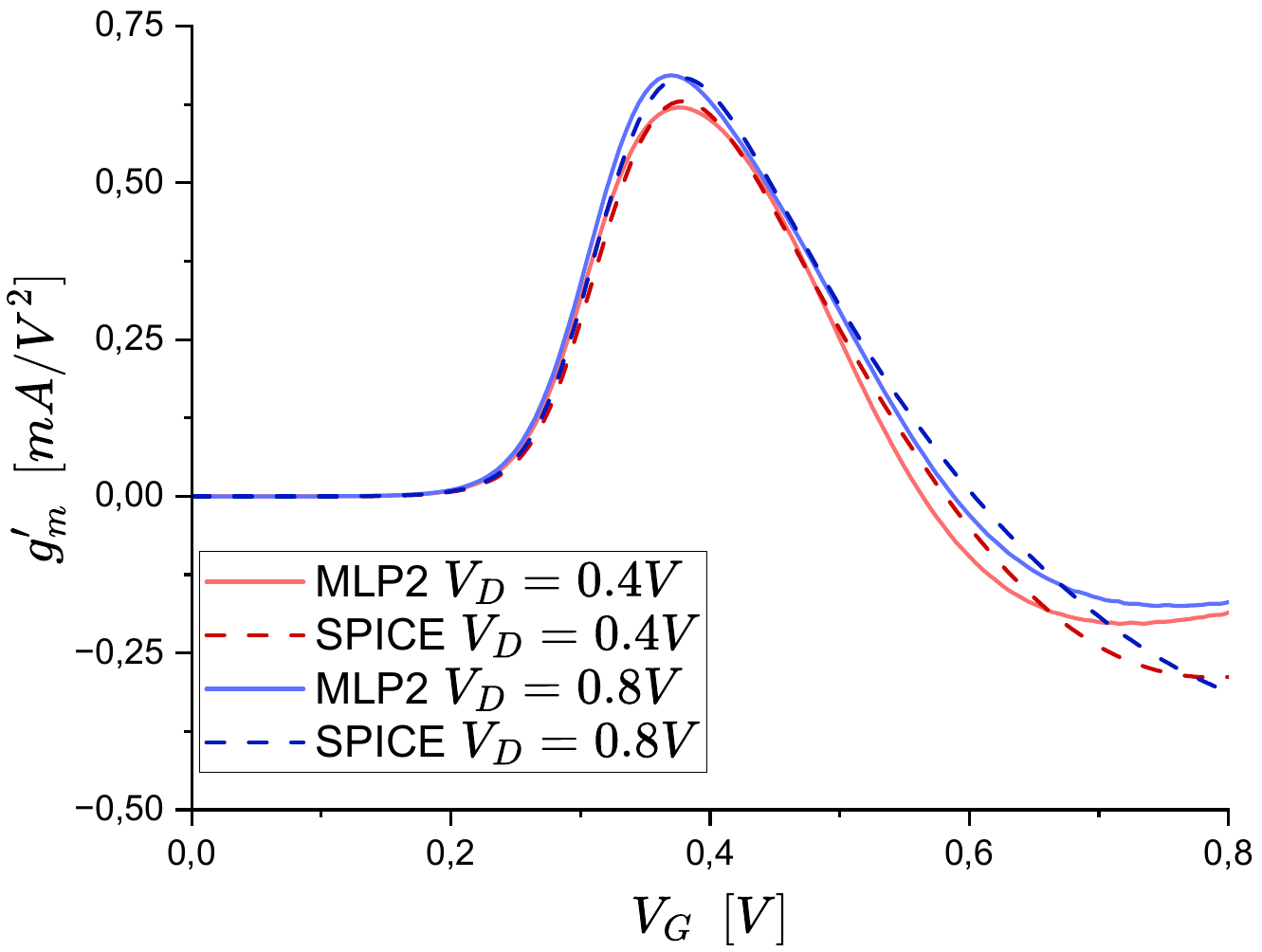}
    \label{fig:derivative_plots_d2_mlp}
\end{subfigure}\hspace{0.01em}%
\begin{subfigure}[b]{0.33\textwidth} 
    \centering
    \caption{$\partial g_m/ \partial V_g$ for KAN2.} 
    \includegraphics[width=\linewidth]{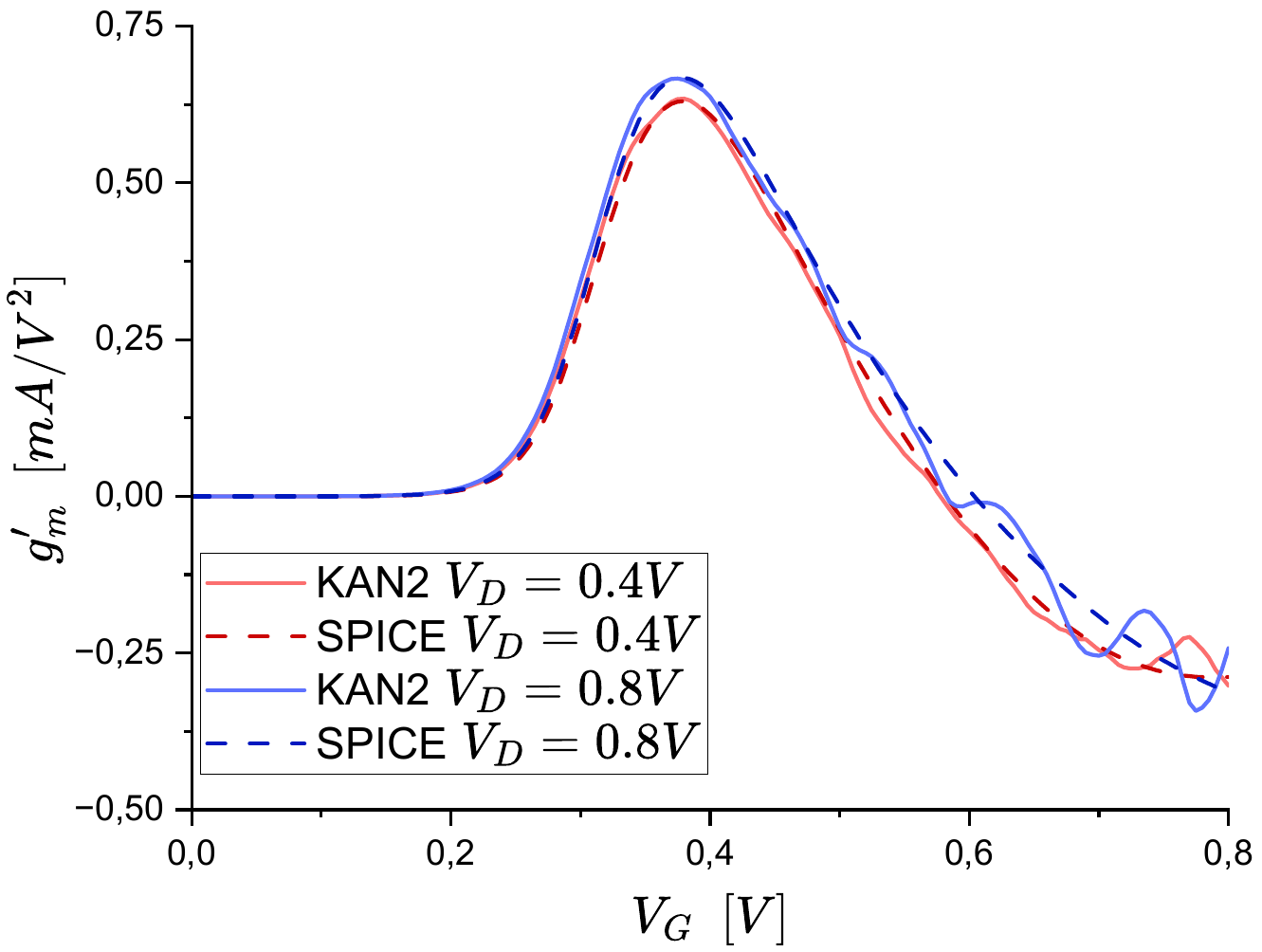}
    \label{fig:derivative_plots_d2_kan}
\end{subfigure}\hspace{0.01em}%
\begin{subfigure}[b]{0.33\textwidth} 
    \centering
    \caption{$\partial g_m/ \partial V_g$ for FKAN2.} 
    \includegraphics[width=\linewidth]{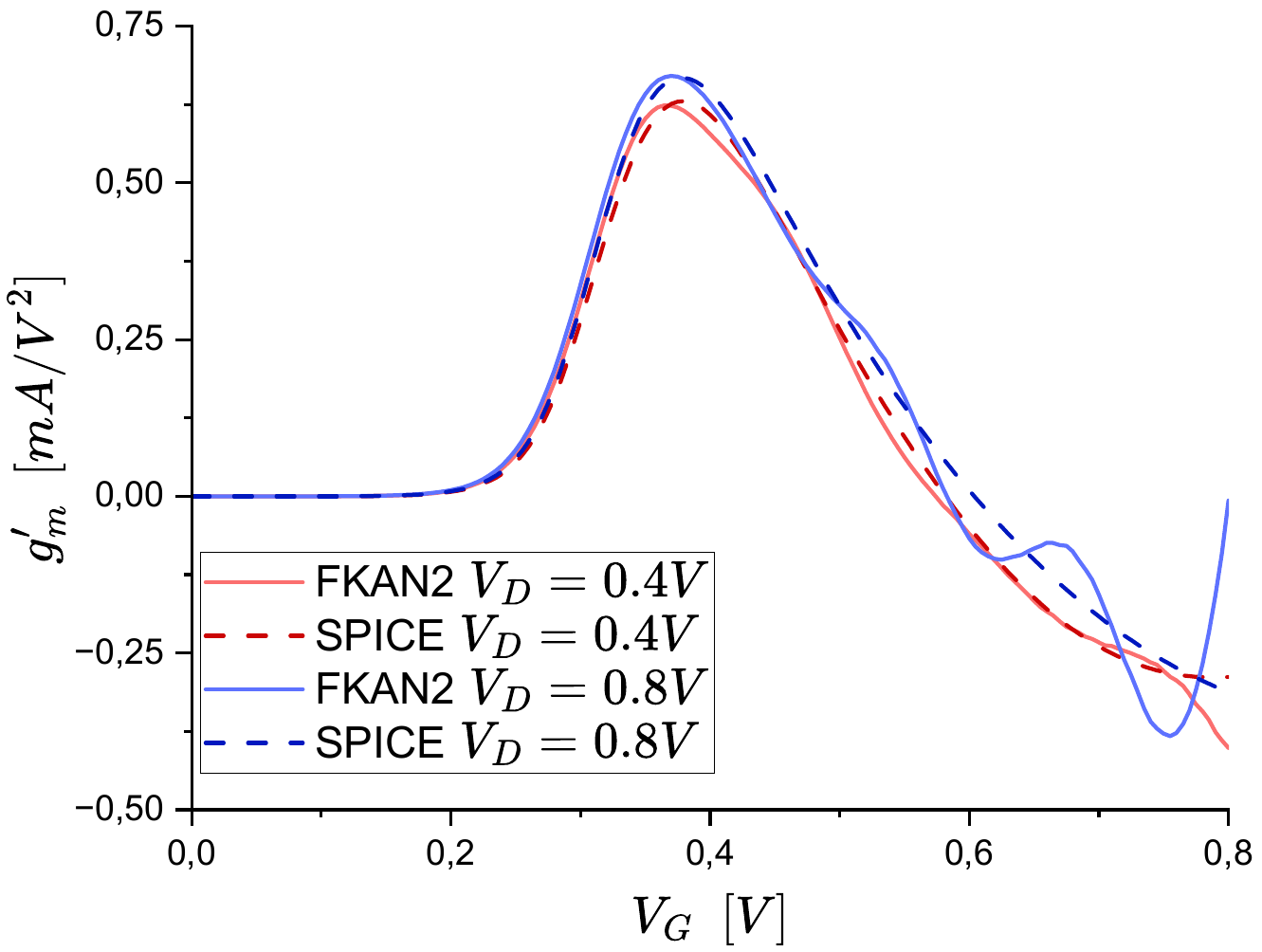}
    \label{fig:derivative_plots_d2_fkan}
\end{subfigure}
    \caption{Transconductance (top) and its derivative (bottom) for networks MLP2, KAN2, and FKAN2 compared against the industry-standard compact model simulated in SPICE (dashed lines). Network names are architectures from Table \ref{table:net_overview}. Since transistors are usually biased around the peak $g_m$ region for higher gain, the slight deviation in the higher-voltage region for \mlp is less important. However, due to inherent properties, \kan and \fkan demonstrate a faulty derivative behavior, which will affect circuit simulations.} 
\label{fig:derivative_plots}
\end{figure*}

\textbf{\mlp}, a backbone for \nn-based transistor modeling, demonstrates \textit{higher error} compared to \kan and \fkan. Even though increasing the model size leads to higher inconsistency with the training, \mlp still produces a stable result with decent generalization. Additionally, \mlp is rather \textit{flexible} with various conversion functions since its performance is not affected by them.

\textbf{\kan} has the \textit{highest fitting capabilities with the lowest number of parameters}. However, \kan demonstrates poor compatibility with the log-scale conversion function, which gives significant benefits to \ids modeling. In addition to inferior generalization, we've experienced considerable difficulties during \kan training for \ids, resulting in a rather high inconsistency and sometimes "NaN" losses, forcing the seed change since training was not recoverable. Nonetheless, increasing the number of parameters made the \ids network training more consistent. On the other hand, charge modeling was much smoother, and having a bigger \kan improved the convergence even further. Dataset 4 training with the voltage step of $50mV$ was still problematic, but it is an extreme case, mainly aimed at testing the limits of generalization since \nn tasks usually require a substantial amount of train data. While the inference speed of KANs was rather slow, they needed much less training time overall due to a significant difference in the number of epochs. Notably, we also tested MultKAN with one hidden layer for \ids prediction, and it produced an error similar to that of KAN2 while being more stable and consistent. However, despite the difference in hidden layers quantity, the number of network parameters between these two architectures was very similar due to the arity for multiplication operators, which effectively parallelized neurons within MultKAN layers.

\textbf{\fkan} proved to be a solid option for accurate transistor modeling. It \textit{takes the best of both worlds} - MLPs simplicity and excellent descriptive characteristics of KANs. While the error is only slightly higher compared to KANs, the training for FKANs is very streamlined and similar to MLPs. FKANs work reliably with a log-scale conversion function and maintain good generalization capabilities. One of the main downsides of FKANs is their inherently large model size due to all stored sine and cosine coefficients. All KANs require a relatively large grid size $G$ to fit complex data patterns since this parameter greatly affects fitting accuracy. Thus, having a small grid size significantly hinders the performance of KANs with transistor modeling. To perform a fair comparison with respect to model size, we set $G_{HL}=2$ for FKAN2. Notably, setting $G_{HL}=8$ leads to a similar performance to KAN2 but results in 1425 network parameters.

\subsection{Derivatives} \label{sec:derivs}
In the previous section, we demonstrated that KANs and FKANs can achieve \textit{very high transistor modeling accuracy}. \ids, \qd, and \qs modeling is essentially a regression task, where KANs and FKANs excel. However, another important quality of a good transistor model is the ability to produce smooth derivatives that \textit{capture device behavior correctly} since it is essential for circuit simulation. 

For testing, we selected MLP2, KAN2, and FKAN2 \ids networks with the lowest test error and plotted transconductance $g_m=\partial I_d/\partial V_g$ and $g'_m=\partial g_m/ \partial V_g$ for two fixed drain voltages $V_{D1}=0.4V$ and $V_{D2}=0.8V$. Fig. \ref{fig:derivative_plots} illustrates the behavior for all selected networks. The first and the second derivatives for MLP2 in Fig. \ref{fig:derivative_plots_d1_mlp} and Fig. \ref{fig:derivative_plots_d2_mlp} demonstrate a slight deviation from the industry-standard compact model in SPICE due to a prediction error. However, they \textit{resemble the correct behavior}, showing a noticeable divergence only in a saturation region with high $V_g$, which is less important since transistors are usually biased around the peak $g_m$ in practice. One of the methods to eliminate the deviation is to increase the \mlp size or/and substantially extend the voltage range for the train data.

In contrast to MLP2, both KAN2 and FKAN2 demonstrate \textit{significant deviations in the derivative behavior} despite having lower prediction errors. For the transconductance, shown in Fig. \ref{fig:derivative_plots_d1_kan} and Fig. \ref{fig:derivative_plots_d1_fkan}, both networks exhibit slight waviness, which is amplified in $g'_m$ plots in Fig. \ref{fig:derivative_plots_d2_kan} and Fig. \ref{fig:derivative_plots_d2_fkan}. The difference in the derivatives can be explained by the \textit{inherent properties} of KANs and FKANs. MLPs use linear operations and nonlinear functions to approximate the data. KANs' output is essentially a \textit{piecewise function consisting of multiple $k$-rank polynomials} with the number of pieces influenced by the grid size $G$. Similar to KANs, FKANs' output is a \textit{piecewise function of multiple sine and cosine terms}, which becomes extremely noticeable in the second-order derivative of \ids in Fig. \ref{fig:derivative_plots_d2_fkan}. Similar KANs' and FKANs' derivative behavior can be observed for \qd and \qs. One of the ways to reduce waviness in derivatives is to reduce $k$ for KANs and $G$ for both architectures, but prediction error will be increased as a result. Another potential option is to \textit{guide} \kan by fixing certain edge functions, thus removing or diminishing the piecewise nature of these functions.

\subsection{Symbolic regression} \label{sec:sreg}
\begin{figure}%
    \centering
    \subfloat[\centering Trained \kan]{{\includegraphics[width=4.8cm]{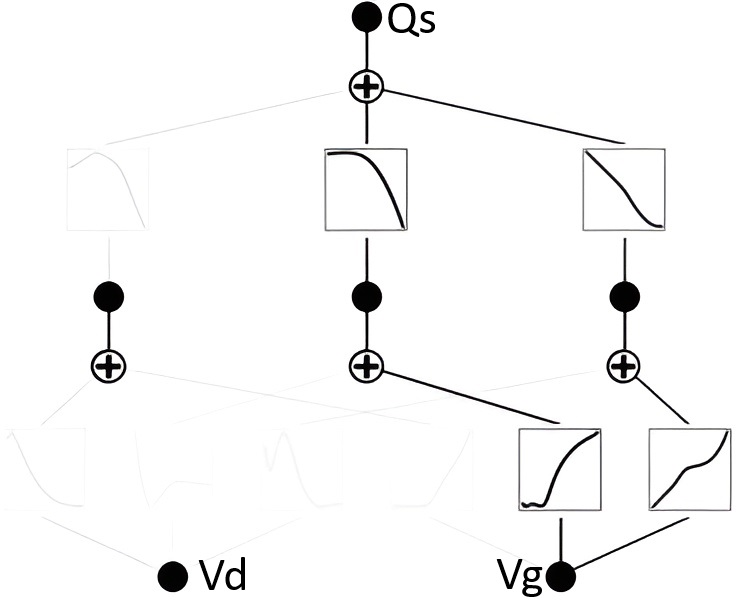} }}%
    \label{fig:symbolic_qs_kan1} 
    \qquad
    \subfloat[\centering Symbolic \kan]{{\includegraphics[width=2.8cm]{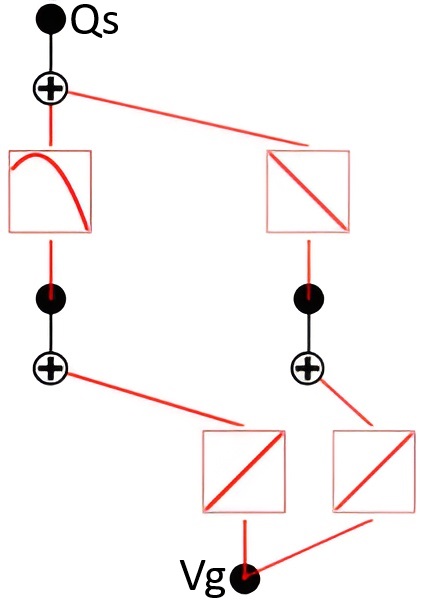} }}%
    \label{fig:symbolic_qs_kan2}
    \caption{\kan symbolic regression for \qs modeling. Plot intensity on the left figure represents the importance of nodes/edges for the output and shows the essential \kan parts for \qs modeling, corresponding to $V_G$.}%
    \vspace{-2ex}
    \label{fig:symbolic_qs_kan}%
\end{figure}

Additionally, we explored the ability of KANs to produce a symbolic formula. At the moment of writing, PyKAN framework generates a symbolic formula by fitting one of the basic functions (e.g. $x^n, 1/x^n, e^x, sin(x)$ and etc.) into each neuron (see Fig. \ref{fig:symbolic_qs_kan}) and combining them into one function based on interconnections between neurons. Thus, the final function, describing the entire \kan (symbolic \kan), could be too convoluted for complex patterns such as transistor \ids, \qd, or \qs curves. Nevertheless, this formula can still be utilized to determine which variable affects the output value the most by comparing the coefficients and variable exponents. As an example, we produced a symbolic formula for \qs KAN1 with the lowest train error:
\begin{equation}\label{eqn:qs_symbolic}
    \begin{split}
    Q_s(V_d,V_g) = -12.83\cdot V_G - 67.91\cdot(V_G - 0.25)^2 + 0.81
    \end{split}
\end{equation}

While Eq. \ref{eqn:qs_symbolic} is only a rough approximation, unusable as a legitimate transistor model, it still managed to capture the fact that \qs does not depend on \vds. A symbolic \kan loses a substantial amount of accuracy since basic functions can rarely fit perfectly into the splines for networks trained on complex data. Fig. \ref{fig:symbolic_qs_kan} illustrates the conversion process and demonstrates the loss of precision in learnable activation functions. Due to inaccurate fitting of basic functions, the resulting \mape of the symbolic KAN shown in Fig \ref{fig:symbolic_qs_kan} and described in Equation \ref{eqn:qs_symbolic} is $91\%$. Thus, KANs \textit{can greatly benefit from the expertise and knowledge of conventional transistor modeling techniques}. In particular, new basic functions could be introduced to guide KANs' training and conversion to a symbolic formula.

\begin{figure}
\centering
\includegraphics[width=1\linewidth]{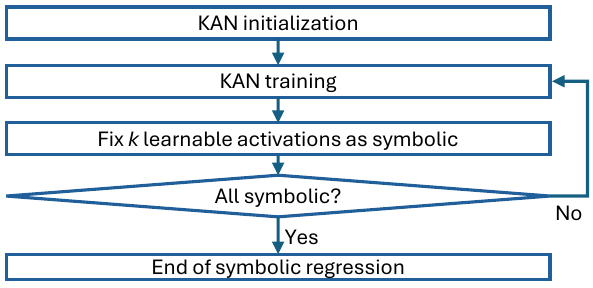}
\caption{Iterative symbolic regression algorithm.}
\label{fig:sr_algorithm}
\end{figure}

\begin{figure}
\centering
\begin{subfigure}[b]{0.45\textwidth}
   \caption{Symbolic KAN for the drain charge \qd shows the importance of both \vds and \vgs for prediction.}
   \includegraphics[width=1\linewidth]{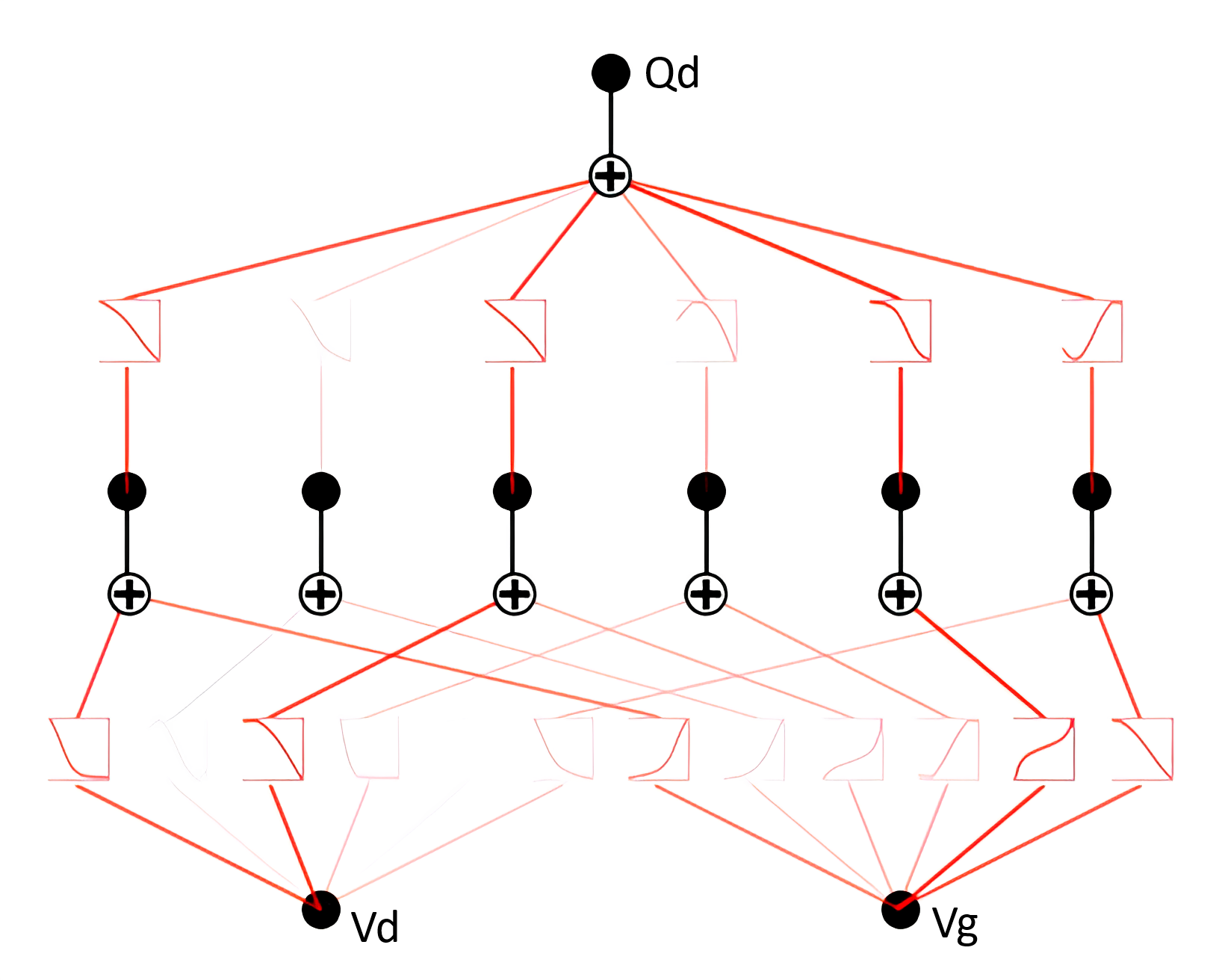}
   \label{fig:qd_isr} 
\end{subfigure}
\begin{subfigure}[b]{0.4\textwidth}
   \caption{Symbolic KAN for the source charge \qs shows the overwhelming influence of \vgs for prediction.}
   \includegraphics[width=1\linewidth]{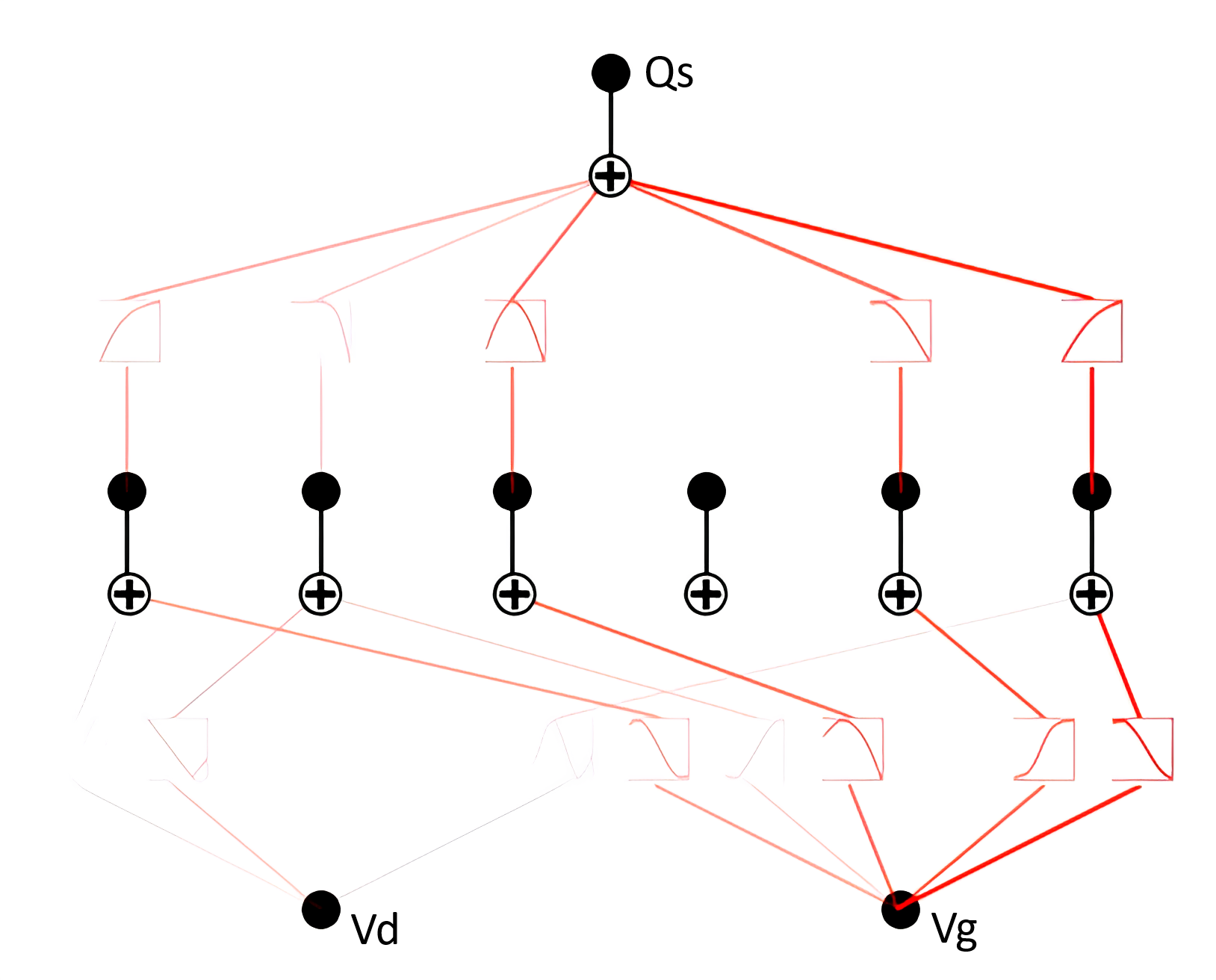}
   \label{fig:qs_isr}
\end{subfigure}
\caption{Symbolic KAN obtained with iterative symbolic regression. Plot intensity indicates the importance of edges or nodes for the output value prediction. All learnable activation functions are fixed and represented with basic functions.}
\label{fig:iterative_sr}
\end{figure}

\subsection{Iterative symbolic regression.} \label{sec:isreg}
The major downside of the post-training symbolic regression demonstrated in Section \ref{sec:sreg} is the substantial accuracy loss due to inability of basic arithmetic functions to correctly represent complex curves produced by learnable activation functions of KANs. While it is possible to find better fitting functions using various tools or using domain knowledge, in this work we would like to explore an alternative solution for symbolic regression, which uses only default basic functions.

\begin{figure*}[!ht]
\centering
\begin{subfigure}[b]{0.48\textwidth} 
    \caption{Inverter.} 
    \includegraphics[width=\linewidth]{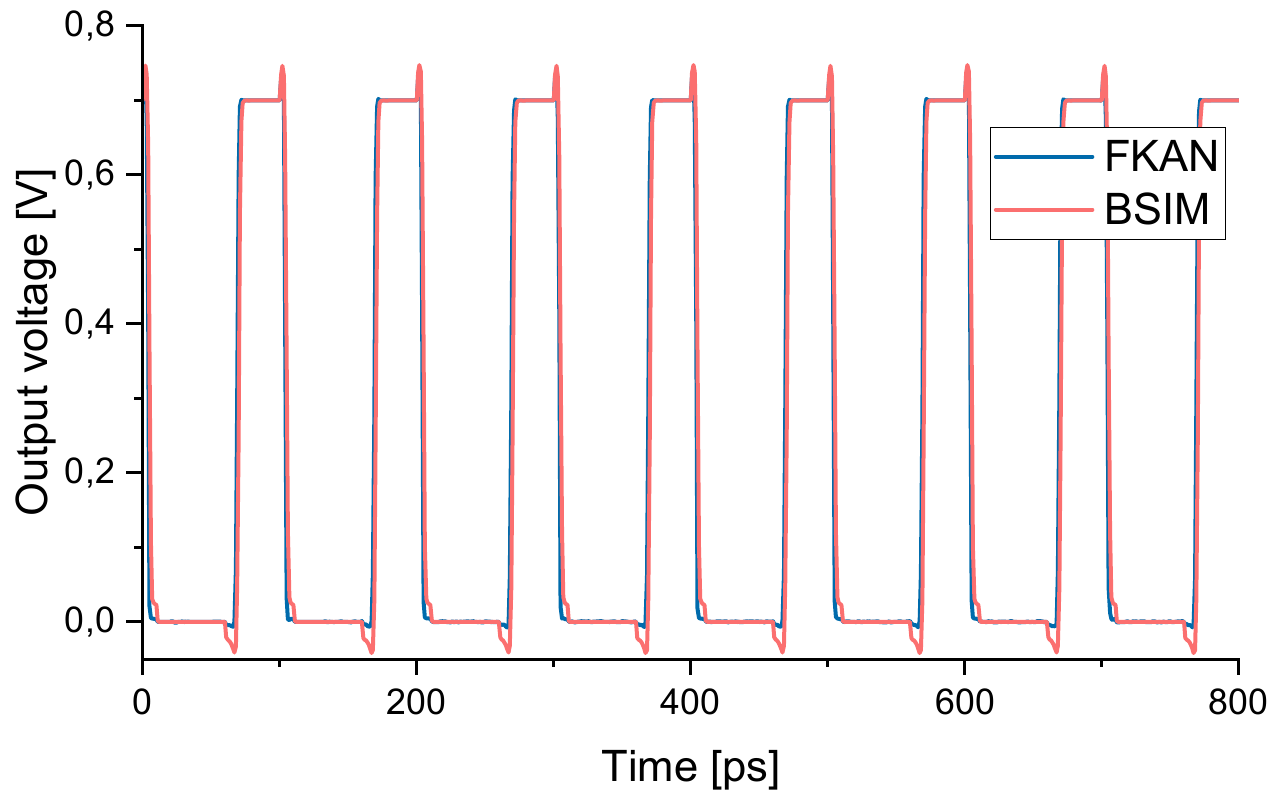}
    \label{fig:spice_inv}
\end{subfigure}\hspace{0.5em}%
\begin{subfigure}[b]{0.48\textwidth} 
    \centering
    \caption{25-stage ring oscillator.} 
    \includegraphics[width=\linewidth]{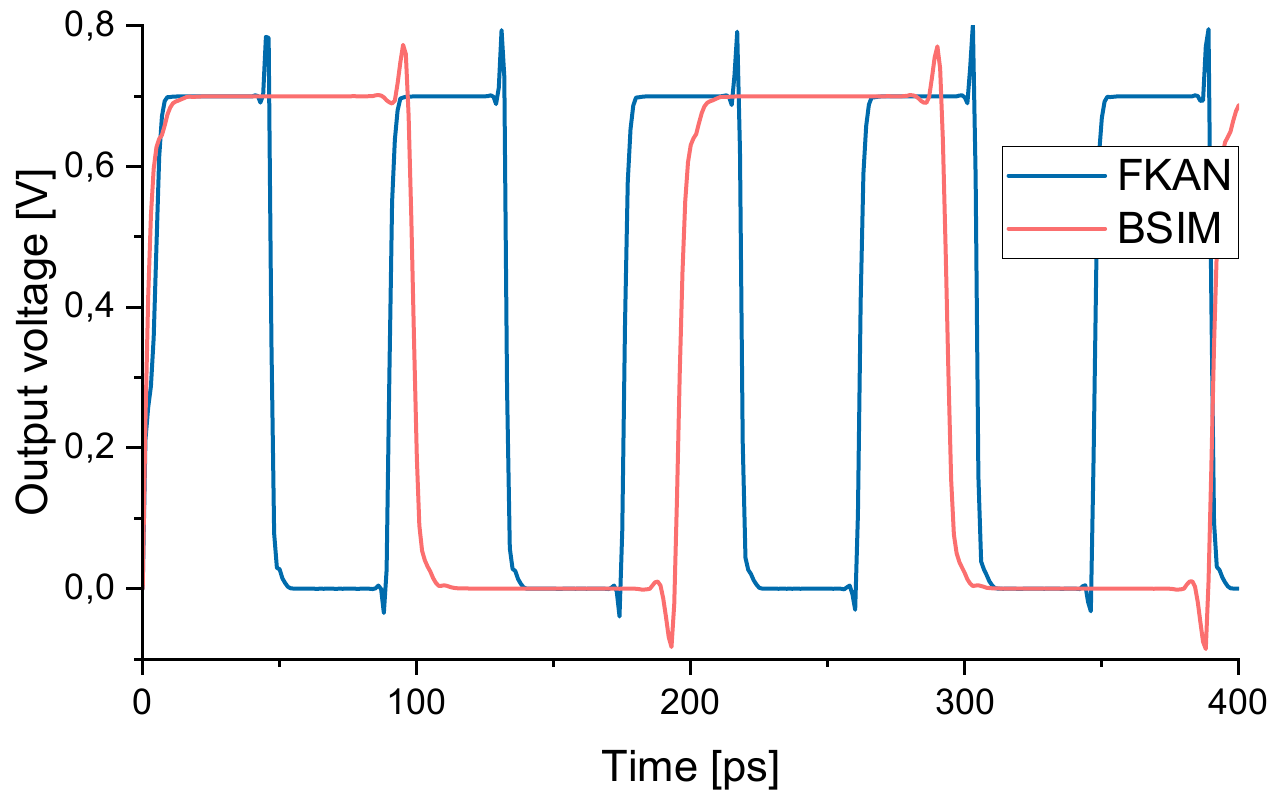}
    \label{fig:spice_osc}
\end{subfigure}
    \caption{SPICE simulations performed for various circuits in \hspice with the industry-standard BSIM-CMG compact model and integrated FKAN transistor model. Inverter simulation demonstrates a good match of FKAN and BSIM models with $<1ps$ front delay. 25-stage ring oscillator illustrates the accumulated delay of 25 inverters due to faulty derivative behavior of FKAN, resulting in a significant difference in frequency between FKAN and BSIM simulation outputs.} 
\label{fig:spice_sim}
\end{figure*}

One of the major restrictions for achieving good symbolic representation for complex functions is the limited number of learnable activation functions in KANs, used in this paper. The original BSIM-CMG transistor model employs a wide range of functions with hundreds of scaling coefficients to achieve high fitting accuracy. Thus, the first step to getting better symbolic regression is to increase the number of learnable functions. The increased number of KAN parameters is irrelevant in this particular case since KAN will be represented as a function similar to Equation \ref{eqn:qs_symbolic} in the end. However, just increasing the number of learnable activation functions will not simplify post-training symbolic regression since KANs tend to lean towards complex curves in these activations. Thus, we propose an \textit{iterative symbolic regression} with a modified PyKAN framework, which aims to improve the fitting accuracy of symbolic KANs.

The algorithm for iterative symbolic regression is shown in Figure \ref{fig:sr_algorithm}. First, KAN is initialized and trained as usual. Next, in the trained KAN $k$ learnable activation functions with the \textit{least accurate} basic function fitting are selected and fixed with the closest basic function. Since the fit was inaccurate, the overall accuracy of KAN will drop significantly even with a handful of changed (fixed) activation functions. The idea behind fixing the least accurate activation functions early is to allow KAN to regain the accuracy with further training. After fixing $k$ activations, KAN is retrained, and another $k$ least accurate activations are fixed. The process is repeated until all activations are fixed with a function, which allows the conversion of the entire KAN into a symbolic form. In practice, after the first training, activation functions are usually fixed with basic functions and R2-score up to 0.6, while closer to the end, R2-scores for fixed functions fluctuate around 0.99, which indicates that KAN is able to regain accuracy after early inaccurate fixes.

To test the algorithm, we selected the KAN configuration with $n_0=2, n_1=6, n_2=1$ resulting in 18 total learnable activation functions. The number of fixed functions after each training iteration was selected as $k=3$ to give KAN sufficient time to adapt to changes and regain accuracy. Resulting symbolic KANs for \qd and \qs after six training iterations are shown in Fig. \ref{fig:iterative_sr} where plot intensity indicates the importance of corresponding edges or nodes for the output value prediction, meaning that the output value is not influenced significantly by nodes or edges which are barely visible. Fig. \ref{fig:qd_isr} represents a symbolic KAN for the drain charge \qd with $1.8\%$ \mape and demonstrates the importance of both \vds and \vgs for the prediction. On the other hand, Fig. \ref{fig:qs_isr} with a symbolic KAN for the source charge \qs prediction highlights the overwhelming influence of \vds for an accurate \qs modeling. The \mape of the symbolic KAN for \qs is $0.96\%$. Since it is unfair to compare these symbolic KANs' accuracy to the one from Section \ref{sec:sreg} due to a significant difference in the number of learnable activation functions, we perform the comparison with the same KANs after the very first training iteration. Performing post-training symbolic regression on \qd and \qs KANs gives $41\%$ and $7.8\%$ \mape correspondingly. Thus, an iterative symbolic regression can greatly increase the accuracy of a symbolic KAN.

All symbolic KANs from Fig. \ref{fig:iterative_sr} can be converted to their respective symbolic formulas. Since these symbolic formulas include all 18 fixed activation functions, they are relatively big and will not be reported here. However, they can be found in the supplementary materials. Notably, \qs formula still contains \vds terms for \qs since the symbolic formula was produced for all nodes and edges regardless of their influence on the output. Omitting all \vds terms in \qs equation (fixing them to zero) gives $8.2\%$ \mape for \qs prediction, which indicates little importance of \vds on the \qs prediction in this particular case since omitting \vgs causes \mape to increase up to $98\%$. While we know that \vds should have no influence on \qs, KAN learns patterns in a slightly different way, and \vds term in KAN will give barely noticeable fluctuations to the learned \qs curve, which can cause some influence on the resulting accuracy. Thus, it is advised that available domain knowledge be applied to ease the KAN training and symbolic regression. For comparison, omitting any of the voltage terms in \qd equation results in over $350\%$ \mape in any case.

In this section, we demonstrated how symbolic regression with KANs can help produce a symbolic formula for an unknown transistor characteristic and determine important variables. However, as mentioned in Section \ref{sec:sreg}, available domain knowledge can greatly increase KAN and symbolic KAN regression capabilities, simplify the task, and potentially focus on terms and dependencies with unknown influence or behavior.

\subsection{Circuit simulation with \fkan.} \label{sec:hspice}
Despite the faulty derivative behavior, observed in Section \ref{sec:derivs} for KANs and FKANs, we still test FKAN performance in circuit simulation. Since both NN architectures exhibit similar faulty derivative behavior, FKAN was chosen due to its high compatibility with the exponential conversion function for \ids shown in Equation \ref{eqn:cf1}. Since SPICE simulations can employ arbitrary values for \vds and \vgs it is crucial to employ NN with good generalization and high test accuracy. Since Verilog-A code does not perform well with loop-heavy NNs \cite{Tung2022, Tung2023} and requires loop unrolling into direct multiplication \cite{Tung24spice}, we code our best-performing FKAN2 models into C code. Since gate charge \qg and bulk charge \qb are also necessary for circuit simulation, we train FKAN2 architecture for \qg reaching $0.16\%$ \mape. Then, \qb is calculated during SPICE simulation as:
\begin{equation}\label{eqn:qb}
    Q_B=-(Q_D+Q_S+Q_G)
\end{equation}
\hspice was selected as a simulation tool as one of the most common ones for circuit and transistor analysis. First, we tested a simple inverter with a periodic voltage pulse. The output voltage comparison of FKAN and regular BSIM-CMG simulation is shown in Fig. \ref{fig:spice_inv}. While there is a minor difference during switching showing a simplified behavior of FKAN, the overall curve matches well with the industry-standard BSIM-CMG compact model. However, there is a time delay of $<1ps$ between the fronts, which is not noticeable in Fig. \ref{fig:spice_inv}. Thus, we additionally test a 25-stage ring oscillator with sequentially connected inverters. Fig. \ref{fig:spice_osc} illustrates a noticeable difference in oscillating frequency, since the minor delay observed in the inverter before accumulates over 25 inverters. Even though we did not experience any conversion issues, the faulty derivative behavior of FKAN significantly impacted the results of the circuit simulation.

\section{Conclusion} \label{sec:concl}

In this work, we explored the novel \kan and \fkan architectures for transistor compact modeling, providing a detailed analysis of their advantages and limitations compared to state of the art. Our results demonstrated that \kan and \fkan offer a superior alternative for variability analysis, achieving significantly lower prediction errors. Specifically, \kan-based transistor models exhibited exceptional accuracy, with only $0.09\%$ \ids error and $0.03\%$ charge error relative to the industry-standard compact model.

However, we identified challenges inherent to \kan and \fkan architectures. Their limited ability to capture derivative behavior impact their accuracy in complex circuit simulation, a critical aspect of transistor modeling. Moreover, we also demonstrated how the symbolic regression capabilities of KANs represent a promising direction.
Despite the existing challenges, we believe KANs hold significant potential to evolve into a powerful modeling framework, particularly when paired with expert-guided refinements. 
This work underscores the transformative potential of \kan-based architectures, paving the way for future advancements in interpretable and accurate transistor modeling—a necessity that becomes increasingly critical as technology rapidly scales into the Angstrom era.

%

   


\begin{IEEEbiography}[{\includegraphics[width=1in,height=1.25in,clip,keepaspectratio]{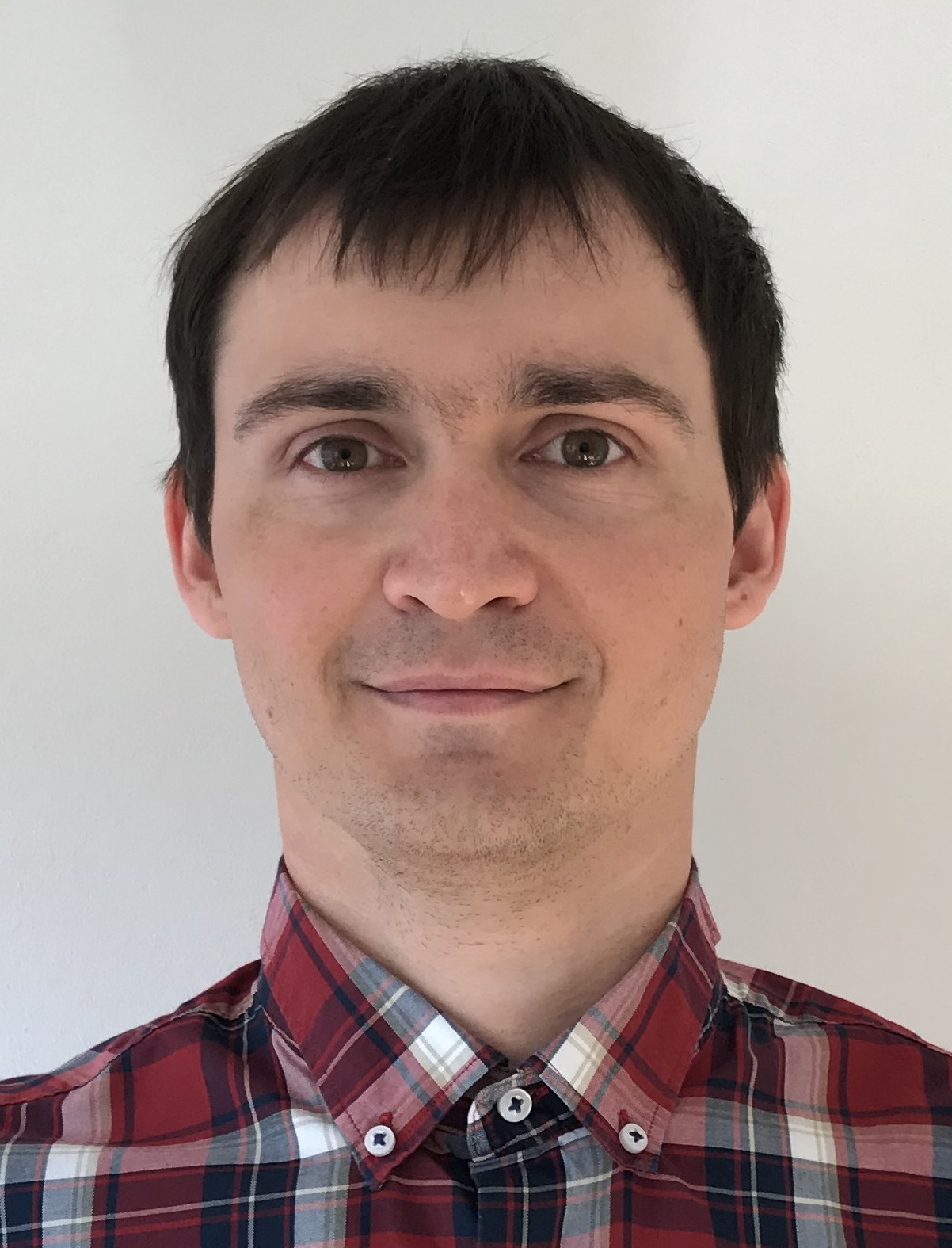}}] {Rodion Novkin} is a Doctoral Researcher at the Chair of AI Processor Design within the Technical University of Munich (TUM). He is also with the Munich Institute of Robotics and Machine Intelligence in Germany. He received the B.Sc. in Robotics and Mechatronics from the Bauman Moscow State Technical University, Russian Federation, in 2013 and the M.Sc. in Automation and Robotics from the Technical University Dortmund, Germany, in 2021. He is currently working towards the Ph.D. degree at the Chair of AI Processor Design within TUM. His research interests include low-power deep and graph neural networks, machine learning for embedded systems and computer-aided design.
\end{IEEEbiography}
\begin{IEEEbiography}[{\includegraphics[width=1in,height=1.25in,clip,keepaspectratio]{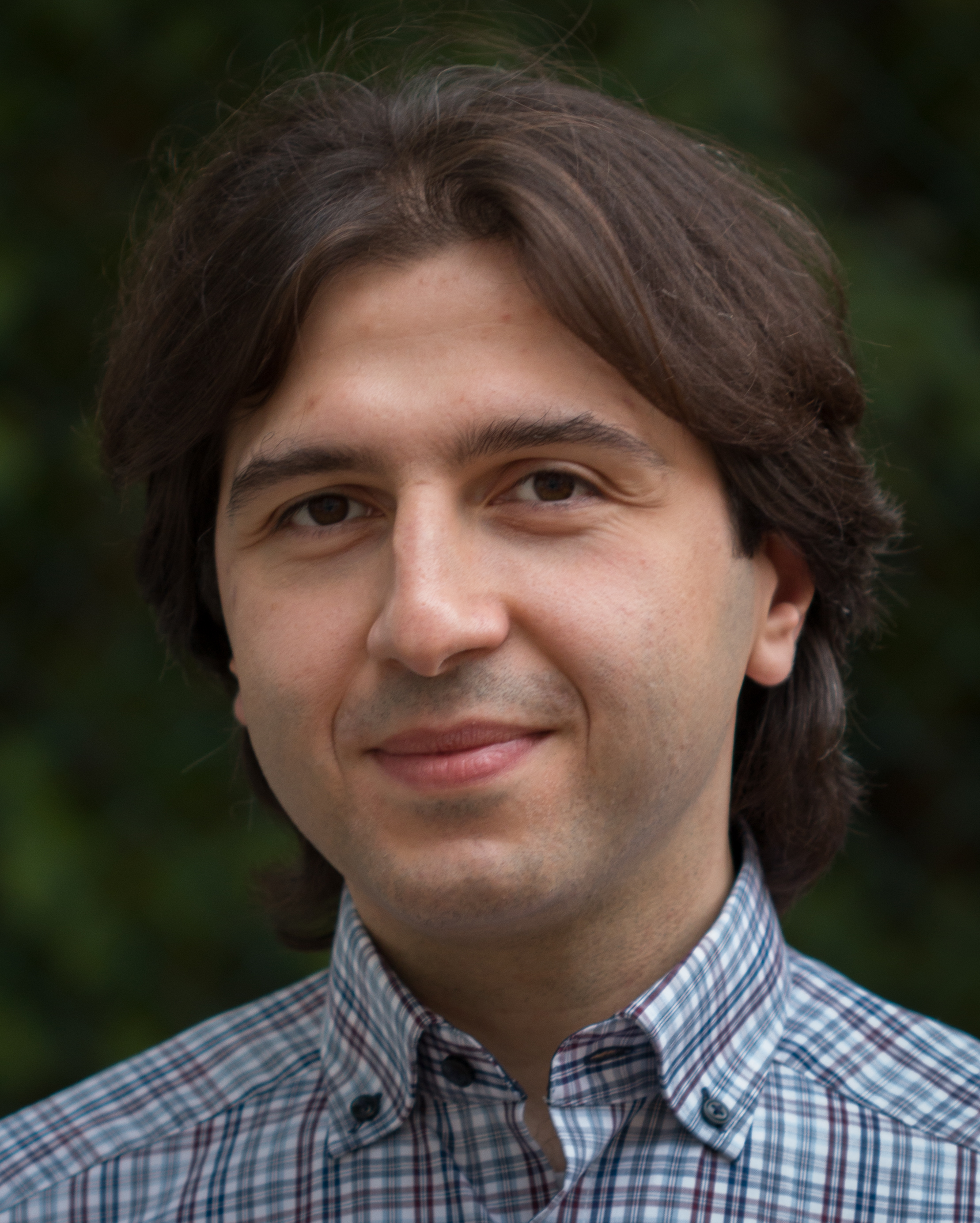}}] {Hussam Amrouch}(S'11-M'15) is Professor heading the Chair of AI Processor Design within the Technical University of Munich (TUM). He is, additionally, heading the Brain-inspired Computing within the Munich Institute of Robotics and Machine Intelligence in Germany and is also the head of the Semiconductor Test and Reliability  within the University of Stuttgart, Germany. He received his Ph.D. degree with the highest distinction (Summa cum laude) from KIT in 2015. His main research interests are design for reliability and testing from device physics to systems, machine learning for CAD, HW security, approximate computing, and emerging technologies with a special focus on ferroelectric devices. He holds 10x HiPEAC Paper Awards and three best paper nominations at top EDA conferences: DAC'16, DAC'17 and DATE'17 for his work on reliability. He has more than $280$ publications in multidisciplinary research areas across the entire computing stack, starting from semiconductor physics to circuit design all the way up to computer-aided design and computer architecture.
\end{IEEEbiography}

\clearpage
\appendices
\section*{Appendix}
\subsection{Symbolic formulas for \qd and \qs corresponding to symbolic KANs in Fig. \ref{fig:iterative_sr}}

\begin{equation}\label{eqn:qs_symbolic_isr}
    \begin{split}
    Q_s(V_d,V_g) = 3.7136\cdot sin(1.0203\cdot tanh(5.8254\cdot V_g - \\ - 2.1197) - 0.0093\cdot atan(8.9127\cdot V_d - 3.4664) + 8.5902) - \\ - 6.1059\cdot cos(8.0838\cdot (0.2769 - V_g)^2 + \\ + 0.0074\cdot cos(8.0775\cdot V_d + 1.3087)- \\ - 4.4072) - 0.0998\cdot cos(0.0191\cdot sin(6.8058\cdot V_d - \\ - 6.7809) + 0.0031\cdot tanh(9.9993\cdot V_g - 4.6827) - 0.2475) + \\ + 23.3674\cdot tanh(-0.0162\cdot sin(8.0467\cdot V_d - 3.6509) + \\ + 0.7524\cdot sin(4.0903\cdot V_g + 1.1983) + 0.7908) + \\ + 5.7207\cdot tanh(0.1042\cdot sin(9.33\cdot V_d + 8.0102) - \\ - 0.6937\cdot sin(4.8929\cdot V_g + 4.1763) + \\ + 1.4648) + 2.182\cdot tanh(1.0677\cdot cos(4.5883\cdot V_g - 6.7694) - \\ - 0.6096\cdot Abs(9.8711\cdot V_d - 6.1893) + 4.6506) - 37.2356
    \end{split}
\end{equation}
\vspace{-13mm}
\begin{equation}\label{eqn:qd_symbolic_isr}
    \begin{split}
    Q_d(V_d,V_g) = -1.4637\cdot (0.7696\cdot sin(4.7539\cdot V_g - 5.1989)+ \\ + tanh(8.2768\cdot V_d - 0.28) - 0.7737)^2 + \\ + 2.5574\cdot sin(2.0392\cdot sin(2.9599\cdot V_g - 1.5741) + \\ + 0.3732\cdot atan(6.1032\cdot V_d - 0.8359) - 2.7426) - \\ - 12.3615\cdot tan(-0.3922\cdot sin(2.5482\cdot V_d + 4.4412) + \\ + 0.0201\cdot tan(3.0281\cdot V_g - 1.0987) + 0.2978) + \\ + 6.5722\cdot tanh(-0.6258\cdot tan(3.1225\cdot V_g + 8.1286) + \\ + 0.0362\cdot tanh(9.9414\cdot V_d - 8.0908) + 0.4307) - \\ - 0.3844\cdot atan(0.4832\cdot sin(5.7174\cdot V_d + 0.7915) + \\ + 2.6052\cdot tanh(3.589\cdot V_g - 2.5902) + 1.545) + \\ + 5.9585\cdot atan(0.8236\cdot tanh(5.5099\cdot V_d - \\ - 0.8742) - 0.4229\cdot atan(6.9059\cdot V_g - \\ - 4.67240142822266) + 0.0923) - 0.5843
    \end{split}
\end{equation}

\end{document}